%% file: ms.tex
\documentclass{article}

     \PassOptionsToPackage{numbers, compress}{natbib}

 \usepackage[nonatbib, preprint]{neurips_2020}

\usepackage[utf8]{inputenc} 
\usepackage[T1]{fontenc}    
\usepackage{hyperref}       
\usepackage{url}            
\usepackage{booktabs}       
\usepackage{amsfonts}       
\usepackage{nicefrac}       
\usepackage{microtype}      

\input{./macro_wnchen.tex}

\input{./preamble_wnchen.tex}

\usepackage{wrapfig, blindtext}

\title{Breaking the Communication-Privacy-Accuracy Trilemma}

\author{%
   Wei-Ning Chen \\
   Department of Electrical Engineering \\
   Stanford University \\
   \texttt{wnchen@stanford.edu} \\
   \And
   Peter Kairouz \\
   Google \\
   \texttt{kairouz@google.com} \\
   \And
   Ayfer \"Ozg\"ur \\
   Department of Electrical Engineering \\
   Stanford University \\
   \texttt{aozgur@stanford.edu} \\
}

\begin{document}

\maketitle
\begin{abstract}
    Two major challenges in distributed learning and estimation are 1) preserving the privacy of the local samples; and 2) communicating them efficiently to a central server, while achieving high accuracy for the end-to-end task. While there has been significant interest in addressing each of these challenges separately in the recent literature, treatments that simultaneously address both challenges are still largely missing. In this paper, we develop novel encoding and decoding mechanisms that simultaneously achieve optimal privacy and communication efficiency in various canonical settings.
    In particular, we consider the problems of mean estimation and frequency estimation under $\varepsilon$-local differential privacy and $b$-bit communication constraints. For mean estimation, we propose a scheme based on Kashin's representation and random sampling, with order-optimal estimation error under both constraints. For frequency estimation, we present a mechanism that leverages the recursive structure of Walsh-Hadamard matrices and achieves order-optimal estimation error for \emph{all} privacy levels and communication budgets. As a by-product, we also construct a distribution estimation mechanism that is rate-optimal for all privacy regimes and communication constraints, extending recent work that is limited to $b=1$ and $\varepsilon=O(1)$. Our results demonstrate that intelligent encoding under joint privacy and communication constraints can yield a performance that matches the optimal accuracy achievable under either constraint alone.

\end{abstract}

\section{Introduction}\label{sec:intro}
\input{sec_introduction.tex}

\subsection{Relation to Prior Work}\label{sec:prelim}
\input{sec_prior_works.tex}


\section{Mean Estimation}\label{sec:mean_est}
\input{sec_mean_estimation.tex}

\section{Frequency Estimation}\label{sec:emp_est}
\input{sec_freq_estimation.tex}

\section{Experiments}\label{sec:exp}
\input{sec_experiments}

\section{Conclusion}\label{sec:conclusion}
\input{sec_conclusion.tex}

\section{Acknowledgments}
The authors would like to thank Jakub Konečný for bringing Kashin's representation to their attention. This was helpful in achieving order-optimality for mean estimation. The authors would also like to thank Vitaly Feldman and Kunal Talwar for pointing out a mistake in the experiments of mean estimation as well as the connection between SQKR and \cite{feldman2017statistical}. This work was supported in part by a Stanford Graduate Fellowship, the National Science Foundation, and a Google Research Award.

\bibliographystyle{abbrv}
\bibliography{reference.bib}

\newpage
\appendix

\input{appendix}



\end{document}

%% file: macro_wnchen.tex
\newcommand{\qedwhite}{\hfill \ensuremath{\Box}}

\newcommand{\lp}{\left(}
\newcommand{\rp}{\right)}
\newcommand{\lb}{\left[}
\newcommand{\rb}{\right]}
\newcommand{\lbp}{\left\{}
\newcommand{\rbp}{\right\}}
\newcommand{\lba}{\left\lvert}
\newcommand{\rba}{\right\rvert}
\newcommand{\lV}{\left\lVert}
\newcommand{\rV}{\right\rVert}
\newcommand{\mv}{\middle\vert}

\newcommand{\mcal}{\mathcal}

\newcommand{\bbm}{\mathbbm}
\newcommand{\mbb}{\mathbb}
\newcommand{\msf}{\mathsf}
\newcommand{\la}{\leftarrow}
\newcommand{\ra}{\rightarrow}

\newcommand{\lan}{\langle}
\newcommand{\ran}{\rangle}

\newcommand{\eqDef}{\triangleq}
\newcommand{\diid}{\overset{\text{i.i.d.}}{\sim}}

\newcommand{\E}{\mathbb{E}}
\newcommand{\Var}{\mathsf{Var}}

\newcommand{\Ber}{\mathrm{Ber}}

%% file: preamble_wnchen.tex
\usepackage{url}
\usepackage{ifthen}
\usepackage{cite}
\usepackage{amsmath} 
\usepackage{subfig}
\usepackage{mathtools}
\usepackage{amsmath,amssymb,mathrsfs}
\usepackage{bm}
\usepackage{bbm}
\newtheorem{lemma}{Lemma}[section]
\newtheorem{theorem}{Theorem}[section]

\newtheorem{corollary}{Corollary}[section]
\newtheorem{definition}{Definition}[section]
\newtheorem{remark}{Remark}[section]
\newtheorem{claim}{Claim}[section]

\usepackage[ruled,vlined]{algorithm2e}
\usepackage[table, dvipsnames]{xcolor}
\usepackage{enumitem}

\newenvironment{proof}[1][Proof]{\textbf{#1.} }{\ \rule{0.5em}{0.5em}}

%% file: sec_introduction.tex
The rapid growth of large-scale datasets has been stimulating interest in and demands for distributed learning and estimation, where datasets are often too large and too sensitive to be stored on a centralized machine. 
When data is distributed across multiple devices, communication cost often becomes a bottleneck of modern machine learning tasks \cite{Feng11}. This is even more so in federated learning type settings, where communication occurs over bandwidth-limited wireless links \cite{kairouz2019advances}. Moreover, as more personal data is entrusted to data aggregators, in many applications it carries sensitive individual information, and hence finding ways to protect individual privacy is of crucial importance. In particular, local differential privacy (LDP) \cite{warner1965randomized, evfimievski2003limiting, dwork2006calibrating, kasiviswanathan2011can} is a widely adopted privacy paradigm, which guarantees that the outcome from a privatization mechanism will not release too much individual information statistically. In this paper, we study the relationship between utility (often in forms of accuracy for certain statistical tasks), privacy, and communication \emph{jointly}.

At first glance, privacy and communication may seem to be in conflict with each other: achieving privacy requires the addition of noise, therefore increasing the entropy of the data and making it less compressible. For instance, consider the mean estimation problem, which appears as a fundamental subroutine in many distributed optimization tasks, e.g. distributed stochastic gradient descent (SGD). Here, the goal is to  estimate the empirical mean of a collection of $d$-dimensional vectors. If we first privatize each vector via \texttt{PrivUnit} in \cite{bhowmick2018protection} (which is optimal under LDP constraints) and then quantize via the \texttt{RandomSampling} quantizer in \cite{g2019vqsgd} (which is optimal under communication constrains), a tedious but straightforward calculation shows that the resulting $\ell_2$ estimation error grows with $d^2$. However, this is far from matching the error rate under each constraint separately, which has a linear dependence on $d$. A similar phenomenon happens in the distribution estimation problem, where each client's data is drawn independently from a discrete distribution $\bm{p}$ with support size $d$. One can satisfy both constraints by first perturbing the data via  the Subset Selection (SS) mechanism \cite{ye2017optimal} (which is optimal under LDP constraints) and then quantizing the noised data to $b$ bits. Again, it can be shown that under such strategy, the $\ell_2$ estimation error of $\bm{p}$ has a quadratic dependence on $d$. This leaves a huge gap to the lower bounds under each constraint separately, which have a linear dependence on $d$. See Section~\ref{sec:separation_is_suboptimal} in the appendix for a detailed discussion.

While there has been significant recent progress on understanding how to achieve optimal accuracy under separate privacy \cite{ye2017optimal, Bassily2015} and communication \cite{zhang2013information, an2016distributed} constraints, as illustrated above a simple concatenated application of these optimal schemes can yield a highly suboptimal performance. Recent works that attempt to break this communication-privacy-accuracy trilemma have been either limited to specific regimes or, as we show, are far from optimal. For example, \cite{acharya2019communication} provides a $1$-bit $\varepsilon$-LDP scheme for distribution estimation which is order-optimal only in the low communication regime ($b=O(1)$) and high privacy regime ($\varepsilon = O(1)$), while \cite{g2019vqsgd} tries to address both constraints in the mean estimation setting, but the error rate achieved under their mechanism is quadratic in $d$ and therefore does not improve on the above baseline. { We note that the general privacy regime (i.e. $\varepsilon = \Omega(1)$) is also of both theoretical and practical interest. For instance, when $n = \Omega\lp d \rp$, one can combine LDP with amplification techniques \cite{erlingsson2019amplification, balle2019privacy, erlingsson2020encode} to ensure stronger central differential privacy.}

This paper closes the above gaps \emph{for any given privacy level $\varepsilon$ and communication budget $b$}. Indeed, our results show that the fundamental trade-offs are determined by the more stringent of the two constraints, and with careful encoding we can satisfy the less stringent constraint \emph{for free}, thus breaking the privacy-communication-accuracy trilemma. For the same privacy level $\varepsilon$, this allows us to achieve the accuracy of existing mechanisms in the literature with drastically smaller communication budget, or equivalently, for the same communication budget achieve higher privacy. It also explains, for example, why $1$-bit communication budget is sufficient under the high privacy regime \cite{acharya2019communication, Bassily2017}. We will demonstrate this phenomenon in various canonical tasks and answer the following question: \emph{``given arbitrary privacy budget $\varepsilon$ and communication budget $b$, what are the fundamental limits for estimation accuracy?''} We next formally define the settings and the problem formulations we consider in this paper.



\subsection{Problem Formulation}

The general distributed statistical tasks we consider in this paper can be formulated as follows: each one of the $n$ clients has local data $X_i \in \mcal{X}$ and sends a message $Y_i \in \mcal{Y}$ to the server, who upon receiving $Y^n$ aims to estimate some pre-specified quantity of $X^n$. Note that $X^n$ are \emph{not necessarily drawn from some distribution}. At client $i$, the message $Y_i$ is generated via some mechanism (a randomized mapping that possibly uses shared randomness across participating clients and the server) denoted by a conditional probability $Q_i(y|X_i)$ satisfying the following constraints.
\paragraph{Local differential privacy} Let $\lp \mcal{Y}, \mcal{B} \rp$ be a measurable space, and $Q(\cdot | x)$ be probability measures for all $x \in \mcal{X}$, with $\lbp Q(\cdot | x) | x\in\mcal{X} \rbp$ dominated by some $\sigma$-finite measure $\mu$ so that the density $Q(y|x)$ exists. A mechanism $Q$ is $\varepsilon$-LDP if
    $$ \forall x, x' \in \mcal{X},\, y \in \mcal{Y}, \, \frac{Q(y|x)}{Q(y|x')} \leq e^\varepsilon. $$
\paragraph{$b$-bit communication constraint} $\mcal{Y}$ satisfies $b$-bit communication constraint if each of its elements can be described by $b$ bits, i.e. $\lba \mcal{Y} \rba \leq 2^b$.

The goal is to jointly design a mechanism (at clients' sides) and an estimator (at the server side) so that the accuracy of estimating some target function $\sum_{i=1}^n f(X_i)$ is maximized. In this paper, we are mainly interested in the \emph{distribution-free} framework, that is, we do not assume any underlying distribution on $X_i$, but we also demonstrate that our results can be extended to probabilistic settings. To this end, we will focus on the following four canonical tasks.

\paragraph{Mean estimation} For real-valued data, we consider the $d$-dimensional unit euclidean ball $\mcal{X} = \mcal{B}_{d}(\bm{0},1)$ and are interested in estimating the \emph{empirical mean} $\bar{X} \eqDef \frac{1}{n}\sum_i X_i$. The goal is to minimize the worst-case $\ell_2$ estimation error defined as
\begin{equation}\label{eq:rme_def}
     r_{\texttt{ME}} \lp \ell_2, \varepsilon, b \rp \eqDef \min_{\lp \hat{X}, Q^n\rp}\max_{X^n \in \mcal{X}^n} \E\lb \left\| \hat{X} - \bar{X} \right\|^2_2 \rb,
\end{equation}
where $Q^n$ satisfies $\varepsilon$-LDP and $b$-bit communication constraints. When the context is clear, we may omit $\varepsilon$ and $b$ in $r_{\texttt{ME}} \lp 
\ell, \varepsilon, b \rp$.  

{
\paragraph{Statistical mean estimation} In the probabilistic version of the mean estimation problem, we assume that $X_i$'s are drawn from some common but unknown distribution $P$ supported on $\bm{B}_d(\bm{0},1)$, the goal is to estimate the \emph{statistical mean} $\theta\lp P \rp = \E_P\lb X_1 \rb$ and to minimize the $\ell_2$ estimation error:
$$ r_{\texttt{SME}} \lp \ell_2, \varepsilon, b \rp \eqDef \min_{\lp \hat{\theta}, Q^n\rp}\max_{X^n \in \mcal{X}^n} \E\lb \left\| \hat{\theta}\lp X^n\rp - \theta\lp P \rp \right\|^2_2 \rb.$$}

\paragraph{Frequency estimation} When $\mcal{X}$ consists of categorical data, i.e. $\mcal{X} = [d] = \lbp 1,...,d \rbp$, we are interested in estimating $D_{X^n}(x) \eqDef \frac{1}{n}\sum_i \bbm{1}_{\lbp X_i = x\rbp}$ for $x \in [d]$. With a slight abuse of notation, $D_{X^n}$ is viewed as a vector $(D_{X^n}(1),...,D_{X^n}(d))$ lying in the $d$-dimensional probability simplex. The worst-case estimation error is defined by
$$ r_{\texttt{FE}} \lp \ell, \varepsilon, b \rp \eqDef \min_{\lp \hat{D}, Q^n\rp}\max_{X^n \in \mcal{X}^n} \E\lb \ell \lp  \hat{D}, D_{X^n} \rp \rb,$$
where $\ell = \lVert \cdot \rVert_\infty, \lVert \cdot \rVert_1,$ or $\lVert \cdot \rVert^2_2$ and again $Q^n$ satisfies $\varepsilon$-LDP and $b$-bit communication constraints. 

\paragraph{Distribution estimation}A closely related setting is that of discrete distribution estimation, where we assume that the $X_i$'s are drawn independently from a discrete distribution $\bm{p}$ on the alphabet $\mcal{X} = [d]$,  and the goal is to estimate $\bm{p}$. In this case, the worst-case error is given by $$ r_{\texttt{DE}}\lp \ell, \varepsilon, b \rp \eqDef \inf_{\lp Q^n, \hat{\bm{p}}\rp}  \sup_{\bm{p} \in \mcal{P}_d} \E \lb \ell(\hat{\bm{p}}, \bm{p}) \rb,$$ where $\mcal{P}_d$ is the $d$-dimensional probability simplex.

We note that these canonical tasks serve as fundamental subroutines in many distributed optimization and learning problems. For instance, the convergence rate of distributed SGD is determined by the $\ell_2$ error of estimating the mean of the local gradient vectors (see \cite{agarwal2018cpsgd} for  more on this connection). Lloyd’s algorithm \cite{lloyd1982least} for k-means clustering or the power-iteration method for PCA can also be reduced to the mean estimation task.

\begin{remark}
In this work, we generally assume the availability of shared randomness across the participating clients and the server. In this case the encoding functions at each node can be explicitly denoted as $Q_i(y|X_i,U)$ where $U$ is a shared random variable that is independent of data, referred to as a public coin. $U$ is also available at the server and the estimator implicitly depends on $U$. In our notation, we suppress this dependence on $U$ for simplicity. The entropy of $U$ is referred as the amount of shared randomness needed by a scheme. In Section~\ref{sec:rand}, we discuss the amount of shared randomness required by our schemes in order to achieve  the optimal estimation error Section~\ref{sec:rand}. We point out that in the statistical settings (i.e. statistical mean estimation and distribution estimation), the optimal estimation error can be achieved without shared randomness.
\end{remark}


%% file: sec_prior_works.tex
\begin{table}[bt!]
\centering
\begin{tabular}{|c|c|c|c|c|}
\hline 
$\vphantom{\Theta \lp \frac{d}{\lp n^2\rp } \rp}$ & Privacy & Comm. & $\ell_2$ error\\ 
\hline 
    SQKR (this work, Thm.~\ref{thm:mean_estimation}) & $\forall\, \varepsilon$ & $\forall\, b $ & $\frac{d}{n\min\lp \varepsilon^2, \varepsilon, b \rp}$ \vphantom{$\lp \frac{d}{n\min\lp \varepsilon^2, \varepsilon, b \rp}\rp$}\\
\hline 
   Cross-polytope\cite{g2019vqsgd} & $\varepsilon \succeq 1$ & $b \succeq \log d$ & $\frac{d^2}{n}$ \vphantom{$\lp \frac{d^2}{n}\rp$}\\
\hline
      Simplex \cite{g2019vqsgd} & $\varepsilon \succeq \log d$ & $b \succeq \log d$ & $\frac{d}{n}$ \vphantom{$\lp \frac{d^2}{n}\rp$}\\
\hline
\end{tabular}
\caption{Comparison between our mean estimation scheme and vqSGD \cite{g2019vqsgd}. 
Our scheme applies to general communication and privacy regimes, and achieves optimal estimation error for all scenarios.} \label{tbl:mean_est}
\end{table}


Previous works in the mean estimation problem \cite{an2016distributed, alistarh17qsgd, wen2017terngrad, wang2018atomo, g2019vqsgd,barnes2020rtopk} mainly focus on reducing communication cost, for instance, by random rotation \cite{an2016distributed} and sparsification \cite{alistarh17qsgd, wen2017terngrad, wangni2018gradient, braverman2016communication}. Among them, \cite{g2019vqsgd} considers LDP simultaneously. It proposes vector quantization and takes privacy into account, developing a scheme for $\varepsilon = \Theta(1)$ and $b = \Theta(\log d)$ with estimation error $O(d^2/n)$. In contrast, the scheme we develop in Theorem~\ref{thm:mean_estimation} achieves an estimation error $O(d/n)$ when $\varepsilon=\Theta(1)$ and $b = \Theta(\log d)$. Moreover, our scheme is applicable for any $\varepsilon$ and $b$ and achieves the optimal estimation error, which we show by proving a matching information theoretic lower bound. See Table~\ref{tbl:mean_est} for a comparison of our results with \cite{g2019vqsgd}. A key step in our scheme is to pre-process the local data via Kashin's representation \cite{lyubarskii2010uncertainty}. While various compression schemes, based on quantization, sparsification and dithering have been proposed in the recent literature and Kashin's representation for communication efficiency \cite{fuchs2011spread, studer2012signal, caldas2018expanding, safaryan2020uncertainty} has been also explored in a few works, it is particularly powerful in the case of joint communication and privacy constraints as it helps spread the information in a vector evenly in every dimension. In \cite{feldman2017statistical}, a similar idea based on Kashin's representation is used to preserve LDP under the context of statistical query models, and although not discussed explicitly in \cite{feldman2017statistical}, it can be further extended to reduce the communication. This helps mitigate the error due to subsequent noise introduced by privatization and compression. 

The recent works of \cite{nguyn2016collecting, wang2019locally} also consider estimating empirical mean under $\varepsilon$-LDP. They show that if the data is from a $d$-dimensional unit $\ell_\infty$ ball, i.e. $X_i \in [-1, 1]^d$, then directly quantizing, sampling and perturbing each entry can achieve optimal $\ell_\infty$ estimation error that matches the LDP lower bound in \cite{duchi2013local}, { where their privatization steps are based on techniques developed in \cite{duchi2013local, bhowmick2018protection}}. Nevertheless, their approach does not yield good $\ell_2$ error in general. Indeed, as in the case of separation schemes discussed in Section~\ref{sec:separation_is_suboptimal}, the $\ell_2$ error of their scheme can grow with $d^2$. We emphasize that in many applications the $\ell_2$ estimation error (i.e. MSE) is a more appropriate measure than $\ell_\infty$. For instance, \cite{agarwal2018cpsgd} shows a direct connection between the MSE in mean estimation and the convergence rate of distributed SGD.

\begin{table}[b!]
\begin{center}
\begin{tabular}{|c|c|c|c|}
\hline 
$\vphantom{\Theta \lp \frac{d}{\lp n^2\rp } \rp}$
& Loss & Estimation error & Communication\\

\hline 
    { Asymmertic RAPPOR \cite{wang2017locally, erlingsson14rappor}}
    & $\ell_2$ 
    & $\Theta \lp\frac{d}{n \min \lp \lp e^\varepsilon -1 \rp^2, e^\varepsilon\rp} \rp$ 
    & $d$ bits \\
\hline 
    RHR (this work, Thm~\ref{thm:emp_dist_estimation})
    & $\ell_2$ 
    & $\Theta\lp \frac{d}{n \min \lp \lp e^\varepsilon -1 \rp^2, e^\varepsilon\rp} \rp$ 
    & $\min\lp\lceil \varepsilon \rceil,\log d\rp$ bits \\
\hline $\vphantom{\Theta \lp \sqrt{\frac{d^2}{n \min{\lp e^\varepsilon, \lp e^\varepsilon-1\rp^2, 2^b\rp}}} \rp}$
     Heavy hitter (Thm.~\ref{thm:emp_dist_estimation} and \cite{Bassily2015})
    & $\ell_\infty$ 
    & $ \Theta \lp\sqrt{\frac{\log d}{n \min \lp \varepsilon, \varepsilon^2 \rp}} \rp$ 
    & $\lceil \varepsilon \rceil$ bits \\    
\hline
\end{tabular}
\caption{Comparison of different frequency estimation schemes.} \label{tbl:emp_est}
\end{center}
\end{table}

Frequency estimation under local differential privacy has been studied in \cite{wang2017locally}, where they propose schemes for estimating the frequency of an individual symbol and minimizing the variance of the estimator. Some of their schemes, while matching the information-theoretic lower bound on $\ell_2$ estimation error under privacy constraints, require large communication. For instance, the scheme Optimal Unary Encoding (OUE){, which can be viewed as an asymmetric version of RAPPOR \cite{erlingsson14rappor},} achieves optimal $\ell_2$ estimation error, but the communication required is $O(d)$ bits, which, as we show in this work, can be reduced to $O(\min(\lceil \varepsilon \rceil,\log d))$ bits. We do this by developing a new scheme for frequency estimation under joint privacy and communication constraints. We establish the optimality of our proposed schemes by deriving matching information theoretic lower bounds on $r_{\texttt{FE}}\lp \ell_2, \varepsilon, b \rp$.

Frequency estimation is also closely related to heavy hitter estimation \cite{hsu12hh, erlingsson14rappor, Bassily2015, zhan16hh, Bassily2017, bun18hh, acharya2019communication}, where the goal is to discover symbols that appear frequently in a given data set and estimate their frequencies. This can be done if the error of estimating the frequency of each individual symbol can be controlled uniformly (i.e. by a common bound), and thus is equivalent to minimizing the $\ell_\infty$ error of estimated frequencies, i.e. $r_{\texttt{FE}}\lp \ell_\infty, \varepsilon, b \rp$. It is shown in \cite{Bassily2015} that in the high privacy regime $\varepsilon = O(1)$, 
$ r_{\texttt{FE}}\lp \ell_\infty, \varepsilon, b \rp = \Theta( \sqrt{\log d /n \varepsilon^2} ),  $
and this rate can be achieved via a $1$-bit public-coin scheme that has a runtime almost linear in $n$ \cite{Bassily2017}. An extension, which we describe in Section~\ref{proof:freq_l_infty} of the appendix, generalizes the achievability in \cite{Bassily2015} to arbitrary  $\varepsilon$ and $b$, achieving
$ r_{\texttt{FE}}\lp \ell_\infty, \varepsilon, b \rp = O(\sqrt{\log d/n \min{\lp\varepsilon^2, \varepsilon, b\rp}}).$
We compare our scheme and existing results in Table~\ref{tbl:emp_est}.

\begin{table}[bt!]
\centering
\begin{tabular}{|c|c|c|}
\hline 
$\vphantom{\Theta \lp \frac{d}{\lp n^2\rp } \rp}$ 
Privacy
& $\varepsilon \in (0, 1)$ & $\varepsilon \in \lp 1, \log d\rp$  \\
\hline $\vphantom{\Theta \lp \frac{d}{n \min{\lp e, \lp e-1\rp^2, 2\rp}} \rp}$
    SS \cite{ye2017optimal} 
    & \cellcolor{OrangeRed!40}$d$ bits 
    & \cellcolor{OrangeRed!40}$ \max\lp \frac{d}{e^\varepsilon}, \log d \rp$
    \\
\hline $\vphantom{\Theta \lp \frac{d}{n \min{\lp e, \lp \varepsilon-1\rp, 2\rp}} \rp}$
    HR\cite{acharya2019hadamard} 
    & \cellcolor{OrangeRed!40}$\log d$ bits 
    & \cellcolor{OrangeRed!40}$\log d$ bits 
    \\
\hline $\vphantom{\Theta \lp \frac{d}{n \min{\lp e, \lp \varepsilon-1\rp, 2\rp}} \rp}$
    $1$bit-HR\cite{acharya2019communication} 
    & \cellcolor{CornflowerBlue!40} $1$ bit
    & -
    \\
\hline $\vphantom{\Theta \lp \frac{d}{n \min{\lp e, \lp \varepsilon-1\rp, 2\rp}} \rp}$
    RHR (this work, Thm.~\ref{thm:dist_estimation})
    & \cellcolor{CornflowerBlue!40}$1$ bit
    & \cellcolor{CornflowerBlue!40}$\min\lp \lceil \varepsilon \rceil,\log d \rp$ 
    \\
\hline
\end{tabular}
    \caption{Comparison between LDP distribution estimation schemes, where blue(or red) color indicates that accuracy of the corresponding scheme is optimal (or not). Under same privacy guarantee, our scheme is more communication efficient while achieves same accuracy.} \label{tbl:dist_compare}
\end{table}

If we further assume $X^n$ are drawn from some discrete distribution $\bm{p}$, then the problem falls into distribution estimation under local differential privacy \cite{duchi2013local, erlingsson14rappor, wang2016mutual, kairouz16, ye2017optimal, acharya2019hadamard, acharya2019communication, acharya2019inference, acharya2019inference2} and limited communication \cite{han2018geometric, zhang2013information, garg2014communication, braverman2016communication, han2018distributed, barnes2019lower, acharya2019inference, acharya2019inference2}. Tight lower bounds are given separately: for instance \cite{ye2017optimal, acharya2019hadamard} shows $r_{\texttt{DE}}\lp \ell_1, \varepsilon, \log d \rp = \Omega(\sqrt{d^2/n\min((e^\varepsilon-1)^2, e^\varepsilon)})$ and \cite{han2018distributed} shows $r_{\texttt{DE}}\lp \ell_1, \infty, b \rp = \Omega(\sqrt{d^2/n 2^b})$.

We show that these lower bounds can be achieved simultaneously (Theorem~\ref{thm:dist_estimation}). Our result recovers the result of \cite{acharya2019communication} when $b=1$ and  $\varepsilon = O(1)$ as a special case. See 
Table~\ref{tbl:dist_compare} for a comparison. 

{ Finally, \cite{Bassily2015} proposes a generic approach to compress the communication of any $\varepsilon$-LDP scheme into $1$ bit by utilizing public randomness. However, this result holds only in the high privacy regime $\varepsilon = O(1)$, and as we show in Section~\ref{sec:rand} it uses much more shared randomness as compared to our schemes. For instance, for mean estimation with $\varepsilon = O(1)$, \cite{Bassily2015} uses $O(d)$ bits of shared randomness, while our scheme SQKR (Theorem~\ref{thm:mean_estimation}) requires only $O(\log d)$ bits to achieve the same performance. Moreover, our schemes extend naturally to statistical settings (i.e. statistical mean estimation and distribution estimation) in which case they do not require shared randomness.}

\subsection{Our Contributions and Techniques}
\label{sec:contributions}
To summarize, our main technical contributions include:
\begin{itemize}[leftmargin=1em,topsep=0pt]
     \item For mean estimation, we characterize the optimal $\ell_2$ error $r_{\texttt{ME}} \lp \ell_2\rp = \Theta\lp d/n\min\lp \varepsilon^2, \varepsilon, b \rp \rp$, by designing a public-coin scheme, Subsampled and Quantized Kashin's Response (SQKR), and proving its optimality by deriving matching information theoretic bounds (in Theorem~\ref{thm:mean_estimation}). Our encoding scheme is based on Kashin's representation \cite{lyubarskii2010uncertainty} and random sampling, which allow the server to construct unbiased estimator of each $X_i$ privately and with little communication. This significantly improves on \cite{g2019vqsgd}, which focuses on the special case $\varepsilon = \Omega(1), b = \log d$ and achieves quadratic dependence on $d$ in that case. 
     
    \item For frequency estimation, we characterize the optimal $\ell_1$ and $\ell_2$ errors under both constraints (in Theorem~\ref{thm:emp_dist_estimation}) and propose an order-optimal public-coin scheme called Recursive Hadamard Response (RHR).
    Our result shows that the accuracy is dominated only by the worst-case constraint, and this implies that one can achieve the less stringent constraint for free. The proposed scheme RHR is based on Hadamard transform, but unlike previous works using Hadamard transform, e.g. \cite{Bassily2017}, we crucially leverage the recursive structure of the Hadamard matrix, which allows us to make the estimation error decay exponentially as $\varepsilon$ and $b$ grow. RHR is computationally efficient, and the decoding complexity is $O(n+d\log d)$. We establish its optimality by showing matching lower bounds on the performance.

    \item We show that RHR easily leads to an optimal scheme for distribution estimation \cite{acharya2019communication, acharya2019hadamard, ye2017optimal}, in which case it does not require shared randomness and achieves order-optimal $\ell_1$ and $\ell_2$ error for all privacy regimes and communication budgets.  We also provide empirical evidence that our scheme requires significantly less communication while achieving the same accuracy and privacy levels as the state-of-the-art approaches. See Section~\ref{sec:exp} for more results.
    
    
\end{itemize}

%% file: sec_mean_estimation.tex
In the mean estimation problem, each client has a $d$-dimensional vector $X_i$ from the Euclidean unit ball, and the goal is to estimate the empirical mean $\bar{X} = \frac{1}{n}\sum_i X_i$ under $\varepsilon$-LDP and $b$ bits communication constraints. This problem has applications in private and communication efficient distributed SGD. The following theorem characterizes the optimal $\ell_2$ estimation error for this setting.

\begin{theorem}\label{thm:mean_estimation}
For mean estimation under $\varepsilon$-LDP and $b$-bit communication constraints, we can achieve 
\begin{equation}\label{eq:mean_est}
    r_{\texttt{\upshape ME}} \lp \ell_2, \varepsilon, b \rp \preceq d/n\min\lp \varepsilon^2, \varepsilon, b \rp.
\end{equation} 

Moreover, if $\min(\varepsilon^2, \varepsilon, b) = o(d)$ and $n\cdot\min(\varepsilon^2, \varepsilon, b) > d$, the above error is optimal.
\end{theorem}

Note that by taking $\varepsilon\rightarrow\infty $ for a fixed $b$, or by taking $b\rightarrow \infty$ for a fixed $\varepsilon$ in part (i), Theorem~\ref{thm:mean_estimation} provides the optimal error when we have the corresponding constraint alone. Furthermore, for finite $\varepsilon$ and $b$ we see that the optimal error is dictated by the error due to one of these constraints, the one that leads to larger error, and hence the less stringent constraint is satisfied for free. This also implies that to achieve the optimal accuracy under $\varepsilon$-LDP constraints, we do not need more than $\lceil  \varepsilon \rceil$ bits. We note that the two conditions for optimality in the theorem are standard and are needed to restrict the problem to the interesting parameter regime.

The lower bounds are obtained by connecting the problem to a specific parametric estimation problem with a distribution supported on the unit ball. The lower bounds $\Omega(\frac{d}{n \varepsilon^2})$ and $\Omega(\frac{d}{n b})$ appear in \cite[Prop.~4]{duchi2013local} and \cite[Thm.~5]{an2016distributed} respectively, and the lower bound $\Omega(\frac{d}{n \varepsilon})$ in Theorem~\ref{thm:mean_estimation} is new. To match this lower bound, we propose a public-coin scheme, Subsampled and Quantized Kashin's Response (SQKR), based on Kashin's representation \cite{lyubarskii2010uncertainty} and random sampling.

\subsection{Subsampled and Quantized Kashin's Response}

For each observation $X_i$, we aim to construct an unbiased estimator $\hat{X}_i$ which is $\varepsilon$-LDP, can be described in $b$ bits, and has small variance. 
Towards this goal, our general strategy is to quantize, subsample, and privatize the data $X_i$. However before this, it is crucial to pre-process each $X_i$ by a carefully designed mechanism to increase the robustness of the signal to noise introduced by sampling and privatization. 

\paragraph{Pre-processing via Kashin's representation} 
We first introduce the idea of a tight frame in Kashin's representation. A tight frame is a set of vectors $\lbp u_j\rbp^N_{j=1} \in \mbb{R}^d$ that satisfy Parseval's identity, i.e.
$ \left\| x \right\|^2_2 = \sum_{j=1}^N \lan u_j, x \ran^2 \, \text{ for all } x \in \mbb{R}^d.$
A frame can be viewed as a generalization of the notion of an orthogonal basis in $\mbb{R}^d$ for $N>d$. 
To increase robustness, we wish the information to be spread evenly across different coefficients. Thus, we say that the expansion 
$ x = \sum_{j=1}^N a_ju_j $
is a Kashin's representation of $x$ at level $K$ if $\max_j \lba a_j \rba \leq \frac{K}{\sqrt{N}}\lV x \rV_2$ \cite{kashin1977section}. \cite{lyubarskii2010uncertainty} shows that if $N>\lp1+\mu\rp d$ for some $\mu>0$, then there exists a tight frame $\lbp u_j \rbp_{j=1}^N$ such that for any $x\in\mbb{R}^d$, one can find a Kashin's representation at level $K = \Theta(1)$. This implies that we can represent each $X_i$ with coefficients $\lbp a_j \rbp_{j=1}^N \in [-c/\sqrt{d}, c/\sqrt{d}]^{c'd}$ for some constants $c$ and $c'$.

\paragraph{Quantization}Each client $i$ computes the Kashin's representation $\lbp a_j \rbp_{j=1}^N \in [-c/\sqrt{d}, c/\sqrt{d}]^{c'd}$ of $X_i$,  and then quantizes each $a_j$ into a $1$-bit message $q_j \in \lbp -c/\sqrt{d}, c/\sqrt{d}\rbp$ with $\E[q_j] = a_j$. This yields an unbiased estimator of $\lbp a_j\rbp_{j=1}^N$, which can be described in $\Theta(d)$ bits in total. Moreover, due to the small range of each $a_j$, the variance of $q_j$ is bounded by $O(1/d)$.
    
\paragraph{Sampling and privatization} To further reduce $\lbp q_j \rbp$ to $k=\min(\lceil \epsilon \rceil, b)$ bits, client $i$ draws $k$ independent samples from $\lbp q_j\rbp_{j=1}^N$ with the help of shared randomness, and privatizes its $k$ bits message via $2^k$-RR mechanism\cite{warner1965randomized, kairouz2016extremal}, yielding the final privatized report of $k$ bits, which it sends to the server.
    
Upon receiving the report from client $i$, the server can construct unbiased estimators $\hat{a}_j$ for each $\lbp a_j\rbp_{j=1}^N$, and hence reconstruct $\hat{X}_i = \sum_{j=1}^N \hat{a}_j u_j$, which yields an unbiased estimator of $X_i$. We show that the variance of $\hat{X}_i$ can be controlled by $O\lp d/\min\lp\varepsilon^2, \varepsilon, b\rp \rp$. Therefore $\frac{1}{n}\sum_i\hat{X}_i$ achieves the order-optimal $\ell_2$ estimation error, establishing the upper bound in Theorem~\ref{thm:mean_estimation}. We provide a detailed description of the scheme and its performance analysis in Section~\ref{sec:mean_est_app}.

{
\begin{remark}\label{rmk:mean_est_rand}
In order to achieve optimal communication efficiency, SQKR uses public randomness at the sampling step. That being said, we can still turn SQKR into a private scheme by using additional communication. See Section~\ref{sec:rand} for more details.
\end{remark}
}

At a high-level, SQKR resembles vqSGD\cite{g2019vqsgd} as both schemes seek a suitably designed representation for $X_i$ before quantizing it. vqSGD represents $X_i$ by a basis $B=\lbp b_1,...,b_K \rbp\subset \mbb{R}^d$ where $B$ is chosen in such a way that its convex hull contains the unit $\ell_2$ ball. Therefore we can write $X_i = \sum_{j=1}^N a_jb_j$ with $\sum_j a_j = 1$. Equivalently, the pre-processing step of vqSGD corresponds to a linear transformation that embeds the $d$-dim $\ell_2$ unit ball into a $N$-dim $\ell_1$ ball. In contrast, Kashin's representation above embeds the $d$-dim $\ell_2$ unit ball into an $N$-dim $\ell_\infty$ ball. Therefore, while both schemes have a pre-processing step of a similar flavor, what is achieved by these steps is quite different. The representation of vqSGD is most efficient when it concentrates the information in a few coefficients, while Kashin's representation spreads the information evenly across different coefficients. The first representation serves us well when we only seek to quantize the signal. 
However, the quantized signal becomes very sensitive to privatization noise. Therefore vqSGD ends up with $O(d^2)$ error in the case of both privacy and communication constraints, while we can achieve $O(d)$ error.



{
\subsection{Application to statistical mean estimation}\label{sec:sme}
For mean estimation, SQKR requires shared randomness so that the server can construct an unbiased estimator. However, for distribution estimation where $X_1,...,X_n \diid P$, we can replace the random sampling with a deterministic partitioning of coordinates among the different clients and circumvent the need for shared randomness. This gives us the following theorem:
\begin{theorem}\label{thm:sme}
   For statistical mean estimation under $\varepsilon$-LDP and $b$ bits communication constraint, we can achieve
    \begin{equation}\label{eq:sme}
    r_{\texttt{\upshape SME}} \lp \ell_2, \varepsilon, b \rp \preceq \frac{d}{n\min\lp \varepsilon^2, \varepsilon, b, d \rp},
\end{equation} 
without shared randomness. Moreover, if $\min(\varepsilon^2, \varepsilon, b) = o(d)$, the above error is optimal (even in the presence of shared randomness).
\end{theorem}

The lower bounds follow from the results of \cite{bhowmick2018protection} (under LDP constraint) and  \cite{zhang2013information} (under communication constraint), and we leave the formal proof of the achievability to Section~\ref{sec:sme_proof}.
}

%% file: sec_freq_estimation.tex
Recall that in the frequency estimation problem, given $X_1,...X_n \in [d]$, we want to estimate the empirical frequency $ D_{X^n}(x)$ under $\varepsilon$-LDP and $b$ bits communication budgets on each $X_i$. The following theorem characterizes the optimal estimation error achievable in this setting.
\begin{theorem}\label{thm:emp_dist_estimation}
For frequency estimation under $\varepsilon$-LDP and $b$ bits communication constraint, we can achieve  
   
(i)  $r_{\texttt{\upshape FE}} \lp \ell_2 \rp
\preceq \frac{d}{n\min{\lbp e^\varepsilon, \lp e^{\varepsilon} -1 \rp^2, 2^b, d\rbp}}, \text{ and }
r_{\texttt{\upshape FE}} \lp \ell_1 \rp
\preceq \frac{d}{\sqrt{n\min{\lbp e^\varepsilon, \lp e^{\varepsilon} -1 \rp^2, 2^b, d\rbp}}};
$

(ii)  $ r_{\texttt{\upshape FE}} \lp \ell_\infty \rp
\preceq \sqrt{\frac{\log d}{ n \min{\lbp \varepsilon^2, \varepsilon, b\rbp}}}.$

Moreover, if $\min\lp e^\varepsilon, \lp e^\varepsilon -1 \rp^2, 2^b \rp = o(d)$ and $n \min\lp e^\varepsilon, \lp e^\varepsilon -1 \rp^2, 2^b \rp \geq d^2$, the errors in (i) are order-optimal. 
\end{theorem}

Note that, similar to Theorem~\ref{thm:mean_estimation}, Theorem~\ref{thm:emp_dist_estimation} shows that for finite $\varepsilon$ and $b$, the error is determined by the error due to one of these constraints, and hence the other less stringent constraint is satisfied for free. It also implies that to achieve the optimal accuracy under $\varepsilon$-LDP constraints, we do not need more than $\min{\lp\lceil  \log_2 e \cdot \varepsilon \rceil, \log d\rp}$ bits.In the rest of the section, we overview the scheme we develop to achieve the optimal error in \eqref{eq:mean_est}.

We next overview the scheme that achieves the error in (i) of Theorem~\ref{thm:emp_dist_estimation}. We call this scheme Recursive Hadamard Response (RHR) as it builds on the recursive structure of the Hadamard matrix. The formal description of the scheme and complete proof of Theorem~\ref{thm:emp_dist_estimation} can be found in Section~\ref{sec:freq_est_app}. 

\subsection{Recursive Hadamard Response }

For notational convenience, we will view $D_{X^n}$ as a $d$-dimensional vector $(D_{X^n}(1),...,D_{X^n}(d))$ and assume $X_i$ is one-hot encoded, i.e. $ X_i = \bm{e}_j$ for some $j\in[d]$, so $D_{X^n} = \frac{1}{n }\sum_i X_i$. 
We further assume, without of loss of generality, that $d = 2^m$ for some $m \in \mbb{N}$. Recall that a Hadamard matrix $H_d\in\{-1,+1\}^{d\times d}$ can be constructed in a recursive fashion as
$$ H_m = 
\begin{bmatrix}
H_{m/2} &H_{m/2}\\
H_{m/2} &-H_{m/2}
\end{bmatrix},$$
where $H_1 = [1]$. 
It can be easily shown that $H_d^{-1} = H_d/d.$ 

Instead of directly estimating $D_{X^n}$, our strategy is to first estimate $H_d\cdot D_{X^n}$ and then perform the inverse transform $H_d^{-1}$ to get an estimate for $D_{X^n}$. So each client will transmit information about $Y_i \eqDef H_d\cdot X_i \in \{-1, 1\}^d$ rather than its original data $X_i$.
\paragraph{The 1-bit case} In this case, each client transmits a uniformly at random chosen entry of  $Y_i$  via any $1$-bit LDP channel (for instance, using the $2$-randomized response (RR) scheme \cite{warner1965randomized, kairouz16, kairouz2016extremal}). Once receiving all the bits of the clients, the server can construct an unbiased estimator of $Y_i$ (since the randomness is public the server knows which entry is chosen for communication by each client). It turns out that this simple $1$-bit scheme achieves optimal $\ell_1$ (and $\ell_2$) error $\Theta(\sqrt{d^2/n\varepsilon^2})$ in the high privacy regime $\varepsilon<1$. This idea is not new and has been used in heavy hitter estimation \cite{Bassily2017} and distribution estimation \cite{acharya2019communication}. However, a key question remains: \emph{how do we minimize the error given an arbitrary communication budget $b$ and privacy level $\varepsilon$?}

\paragraph{Moving beyond the 1-bit case} A natural way to extend the $1$-bit scheme above to the case when each client can transmit $b$-bits is to have each client communicate $b$ randomly chosen entries of its transformed data $Y_i$ instead of a single entry. This will boost the sample size by a factor of $b$, equivalently decrease the $\ell_2$ error by a factor of $b$ ($\sqrt{b}$ for $\ell_1$). Instead, we argue next that we can exploit the recursive structure of the Hadamard matrix to boost the sample size by a factor of $2^b$, equivalently decrease the error by an exponential factor.

Consider $b\leq \lfloor \log d \rfloor$ and let $B = d/2^{b-1}$. Note that $H_d=H_{2^{b-1}} \otimes H_B$, where $\otimes$ denotes the Kronecker product. To visualize, for $b=3$, $H_d$ has the following structure:
$$ Y_i=H_d X_i = 
\begin{bmatrix}
H_{B} & H_{B}& H_{B}& H_{B}\\
H_{B} & -H_{B}& H_{B}& -H_{B}\\
H_{B} & H_{B}& -H_{B}& -H_{B}\\
H_{B} & -H_{B}& -H_{B}& H_{B}
\end{bmatrix}
\begin{bmatrix}
X_i^{(1)}\\
X_i^{(2)}\\
X_i^{(3)}\\
X_i^{(4)}
\end{bmatrix},$$
where for $l=1,\dots, 2^{b-1}$, $X_i^{(l)}$ denotes the $l$'th block of $X_i$ of length $B=d/2^{b-1}$. Therefore, in order to communicate $Y_i$, we can equivalently communicate $H_B X_i^{(l)}$ for $l=1,\dots, 2^{b-1}$. Since $H_{2^{b-1}}$ is known, this is sufficient to reconstruct $Y_i$. We next observe that while communicating $Y_i$ requires $d=B\times 2^{b-1}$ bits, communicating $\{H_B X_i^{(l)}, l=1,\dots, 2^{b-1}\}$ requires $B+(b-1)$ bits. This is because $X_i$ is one-hot encoded and all but one of the $2^{b-1}$ vectors $\{H_B X_i^{(l)}, l=1,\dots, 2^{b-1}\}$ are equal to zero. It suffices to communicate the index $l$ of the non-zero vector, by using $(b-1)$ bits, and its $B$ entries by using additional $B$ bits. This is the key observation that RHR builds on.

When each client has only $b$ bits, they cannot communicate sufficient information for fully reconstructing $Y_i$, i.e. all $\{H_B X_i^{(l)}, l=1,\dots, 2^{b-1}\}$. Instead, each client chooses 
a random index $r_i\in[B]$ and communicates the $r_i$'th row of $\{H_B\,X_i^{(l)}, l=1,\dots, 2^{b-1}\}$, equivalently $\{(H_B)_{r_i} X_i^{(l)}, l=1,\dots, 2^{b-1}\}$ where $(H_B)_{r_i}$ denotes the $r_i$'th row of $H_B$. Note that as before, only one of the $2^{b-1}$ numbers $\{(H_B)_{r_i}\, X_i^{(l)}, l=1,\dots, 2^{b-1}\}$ is non-zero and therefore these numbers  can be communicated by using $b$ bits, $b-1$ bits to represent the index of the non-zero number and a single bit to communicate its value. When there is a privacy constraint, client $i$ perturbs their $b$ bits by a $2^b$-RR mechanism with privacy level $\varepsilon$, and this yields the privatized report of $b$ bits.




Upon receiving the reports from clients, the server constructs an unbiased estimator for $Y_i$. To do this, it first constructs an unbiased estimator for $\{H_B\,X_i^{(l)}, l=1,\dots, 2^{b-1}\}$ and then employs the structure $H_d=H_{2^{b-1}} \otimes H_B$. Note that since the randomness is shared the server knows the index $r$ chosen by each client, and since the clients choose their indices independently and uniformly at random, roughly speaking, they communicate information about different rows of $\{H_B\,X_i^{(l)}, l=1,\dots, 2^{b-1}\}$.
Finally, an unbiased estimator $\hat{Y}_i$ for $Y_i$ yields an unbiased estimator for $X_i$ through the transformation $\hat{X}_i = \frac{1}{d}H_d\cdot\hat{Y}_i$, and due to the orthogonality of $H_d$, it can be shown that the variance of $\hat{X}_i$ is the same as the variance of $\hat{Y}_i$ divided by $d$. 

A subtle issue is that if $e^\varepsilon \ll 2^b$, the noise due to $2^b$-RR mechanism may be too large, so instead of using all $b$ bits, we perform the above encoding and decoding procedure with $b' \eqDef \min\lp \lceil\log_2 e \cdot \varepsilon \rceil \rp$. 
We defer the details and the formal proof to Section~\ref{sec:proof_freq_est_achievability}.

Note that this careful construction based on the recursive structure of the Hadamard matrix is only required in the case when there are joint privacy and communication constraints. When only one constraint is present, the optimal error can be achieved in a much simpler fashion. When there is only a $b$ bit constraint, \cite{han2018distributed} shows that the optimal error can be achieved by simply having each client communicate a subset of the entries of its data vector $X_i$ (without requiring Hadamard transform). When there is only a privacy constraint $\varepsilon$, the optimal error can be achieved by a number of schemes, such as subset selection ($2^b$-SS)\cite{ye2017optimal} and Hadamard response (HR) \cite{acharya2019hadamard}.

The encoding mechanism above involves two operations: 1) sampling a random index $r_i$ from $[B]$ at each client with the help of a public coin, and 2) computing $\lp H_d\rp_{r_i}\cdot X_i$. Since $X_i$ is one-hot, the encoding complexity is $O(\log d)$. On the other hand, in order to efficiently decode, the server first computes the joint histogram of client $i$'s report and $r_i$ in $O(n)$ time, which in turn allows us to calculate $\frac{1}{n}\sum_i \hat{Y}_i$, and then apply the Fast Walsh-Hadamard transform (FWHT) to obtain the estimator of empirical frequency in $O(d\log d)$ time. Hence the overall decoding complexity is $O\lp n+ d\log d \rp$. See Algorithm~\ref{alg:freq_est_encoding} and Algorithm~\ref{alg:freq_est_decoding} in Section~\ref{sec:freq_est_app} for details.

{
\begin{remark}
As in mean estimation, RHR requires public randomness to achieve optimal communication efficiency. Indeed, we can show that RHR uses the minimum amount of shared randomness. See Section~\ref{sec:rand} for more details.
\end{remark}
}

\subsection{Application to distribution estimation}\label{sec:dist_est}

As in statistical mean estimation (Section~\ref{sec:sme}), for distribution estimation where $X_1,...,X_n \diid \bm{p}$, we can replace the random sampling with deterministic one and avoid the use of shared randomness. This yields the following theorem:
\begin{theorem}\label{thm:dist_estimation}
   For distribution estimation under $\varepsilon$-LDP and $b$ bits communication constraint, we can achieve
    \begin{equation*}
        r_{\texttt{\upshape DE}} \lp \ell_2 \rp
        \asymp \frac{d}{ n \min{\lp e^\varepsilon, \lp e^{\varepsilon} -1 \rp^2, 2^b, d\rp}}, \text{ and }
        r_{\texttt{\upshape DE}} \lp \ell_1\rp
        \asymp \frac{d}{\sqrt{n\min{\lp e^\varepsilon, \lp e^{\varepsilon} -1 \rp^2, 2^b, d\rp}}},
    \end{equation*}
    without shared randomness. Moreover, if $n\cdot \min\lp e^\varepsilon, \lp e^\varepsilon -1 \rp^2, 2^b, d \rp \geq d^2$, the above errors are optimal even in the presence of shared randomness.
\end{theorem}

The lower bounds follow directly from the results of \cite{ye2017optimal} (under LDP constraint) and  \cite{han2018distributed, barnes2019lower} (under communication constraint). We leave the formal proof of the achievability to Section~\ref{sec:dist_est_app}.


{
\section{Role of Shared Randomness and How It Benefits Communication}\label{sec:rand}

\paragraph{The Amount of Shared Randomness}
In the achievability part of Theorem~\ref{thm:mean_estimation}, our proposed scheme SQKR randomly and independently samples $b^*_{\texttt{ME}} \eqDef \min\lp \lceil \varepsilon \rceil, b \rp$ bits from the quantized $d$-dimensional binary vector at each client. These bits are then privatized and communicated to the server. In addition to the values of these bits, the server needs to know the indices of the sampled bits, which corresponds to an additional $b^*_{\texttt{ME}}\log d$ bits of information that needs to be shared between each client and the server. This information can be shared in two different ways: 1) sampling can be done by using a public coin shared a priori between the client and the server, or 2) sampling can be done by using a  private coin at the client side, which is then communicated to the server. We can also combine both 1) and 2) when $b>b^*_{\texttt{ME}}$: given $b$ bits communication budget, SQKR compresses the data to $b^*_{\texttt{ME}}$ bits, so the client can use the remaining $b-b^*_{\texttt{ME}}$ bits to communicate the locally generated randomness required at the sampling step. Thus the amount of shared randomness is reduced to $b^*_{\texttt{ME}}\log d - (b-b^*_{\texttt{Me}})$ bits. Moreover, by extending \cite[Theorem~4]{acharya2019communication}, we also obtain a lower bound on the amount of shared randomness required, which we summarize in the following corollary:
\begin{corollary}\label{cor:me_rnd}
Under $\varepsilon$-LDP and $b$-bit communication constraints, SQKR uses $\min\lp b^*_{\texttt{ME}}\log d, d \rp - (b- b^*_{\texttt{ME}})$ bits of shared randomness to achieve $r_{\texttt{ME}}\lp \ell_2, b, \varepsilon \rp$, where $b^*_{\texttt{ME}} \eqDef \min\lp \lceil \varepsilon \rceil, b \rp$. Moreover, if $b < \log d - 2$, any $b$-bit consistent mean estimation scheme\footnote{A scheme is \emph{consistent} if it has vanishing estimation error as $n \ra \infty$.} requires at least $\log d - b -2$ bits.
\end{corollary}

We contrast this with the amount of shared randomness needed in the generic scheme of \cite{Bassily2015} which provides $\varepsilon$-LDP by using $1$ bit per client in the high privacy regime $\varepsilon = O(1)$. The shared randomness required by this scheme is $d$ bits per client. In contrast, when $\varepsilon = O(1)$ and $b=1$, SQKR requires $\log d$ bits of shared randomness.

Similarly, for frequency estimation, it can be seen that RHR requires $\log d - b_{\texttt{FE}}^*$ bits of shared randomness in the random sampling step, where $b_{\texttt{FE}}^* \eqDef \min\lp \lceil \varepsilon \log_2 e \rceil, b \rp$. Again, this is achieved by communicating  $b-b_{\texttt{FE}}^*$ bits of  privately generated randomness from the client to the the server, which reduces the required public randomness to $\log d - b$ bits. Furthermore, as in mean estimation, we can show that at least $\log d - b -2$ bits are needed to get a consistent scheme, so RHR is also optimal in the amount of public randomness it uses. We summarize it in the following corollary:
\begin{corollary}\label{cor:fe_rnd}
Under $\varepsilon$-LDP and $b$-bit communication constraints, RHR uses $\log d- b$ bits of shared randomness to achieve $r_{\texttt{FE}}\lp \ell_2, b, \varepsilon \rp$, where $b^*_{\texttt{FE}} \eqDef \min\lp \lceil \varepsilon \log_2 e \rceil, b \rp$. Moreover, if $b < \log d - 2$, any $b$-bit consistent frequency estimation scheme requires at least $\log d - b -2$ bits of shared randomness. Thus RHR is optimal in the amount of shared randomness it uses for frequency estimation, up to an additive constant.
\end{corollary}

The achievability parts of Corollary~\ref{cor:me_rnd} and Corollary~\ref{cor:fe_rnd} follow directly from the analysis of SQKR and RHR, and we defer the proof of the converse part to Section~\ref{sec:proof_cor_rnd}. Given a $\varepsilon$-LDP constraint, we summarize the minimum amounts of communication and shared randomness required to achieve the optimal error $r_{\texttt{ME}}\lp \ell_2, \varepsilon, \infty \rp$ and $r_{\texttt{FE}}\lp \ell_2, \varepsilon, \infty \rp$ in Table~\ref{tbl:rand}. 

\begin{table}[htbp]
\begin{center}
\begin{tabular}{|c|c|c|}
\hline 
$\vphantom{\Theta \lp \frac{d}{\lp n^2\rp } \rp}$
& Communication & Shared randomness\\

\hline 
    SQKR (Thm.~\ref{thm:mean_estimation}) $\vphantom{\Theta \lp \frac{d}{\lp n^2\rp } \rp}$
    & $\lceil \varepsilon \rceil$ bits 
    & $\min\lp \lceil \varepsilon \rceil\log d, d \rp$ bits 
     \\
\hline 
    RHR (Thm.~\ref{thm:emp_dist_estimation}) $\vphantom{\Theta \lp \frac{d}{\lp n^2\rp } \rp}$
    & $\lceil \log_2 e \cdot \varepsilon \rceil$ bits 
    & $\log d - \lfloor \log_2 e \cdot \varepsilon \rfloor $ bits \\
\hline
\end{tabular}
\caption{The amounts of required shared randomness.} \label{tbl:rand}
\end{center}
\end{table}

In Figure~\ref{fig:pub_rand}, we plot the achievable region for the minimax frequency estimation error under $\varepsilon$-LDP constraint (i.e. $r_{\texttt{FE}}\lp \ell_2, \varepsilon, \infty \rp$). Note that the red line in Figure~\ref{fig:pub_rand} can be achieved by RHR.

\begin{figure}[htbp]
  	\centering

  	\subfloat{{\includegraphics[width=0.7\linewidth]{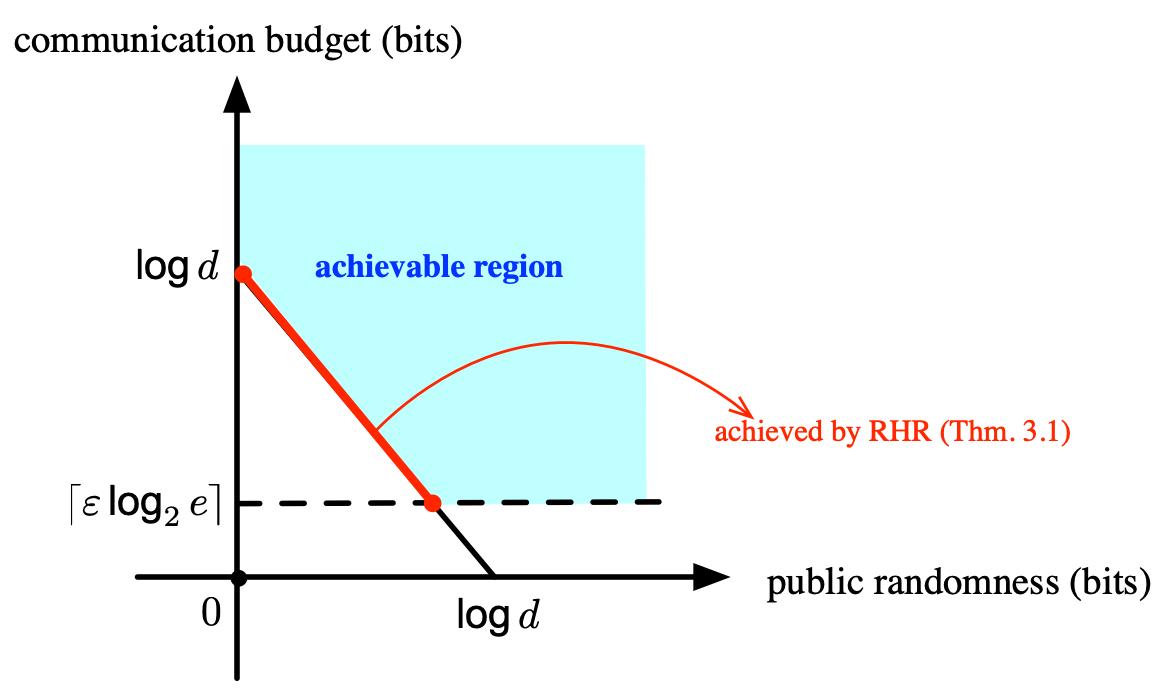} }}%
  	\caption{ Achievable region for frequency estimation with public randomness.}%
  	\label{fig:pub_rand}
\end{figure}

\begin{remark}
Note that shared randomness is only needed for distribution-free settings;  for distribution estimation and statistical mean estimation, one can achieve the same estimation error with only private randomness as noted in Theorems \ref{thm:sme} and \ref{thm:dist_estimation} .
\end{remark}


\paragraph{ Converting public-coin schemes to private-coin schemes}
As discussed above, we can always replace shared randomness with additional communication by first generating the random bits at the client side and then sending them to the server. Therefore, by Corollary~\ref{cor:me_rnd} and Corollary~\ref{cor:fe_rnd}, we automatically obtain private-coin SQKR and private-coin RHR by using additional communication. We next state these observations for completeness. 
\begin{corollary}[Private-coin SQKR]\label{cor:priv_SQKR}
Under $\varepsilon$-LDP and $b$-bit communication constraints with $b > \log d$ and $0<\varepsilon \leq d$, the $\ell_2$ minimax error for private-coin mean estimation, denoted as $\tilde{r}_{\texttt{ME}}(\ell_2, \varepsilon, b)$\footnote{The definition of $\tilde{r}_{\texttt{ME}}(\cdot)$ is the same as that of $r_{\texttt{ME}}(\cdot)$ in \eqref{eq:rme_def}, except that now the minimum is taken over all private-coin schemes.} (to distinguish it from the  minimax error $r_{\texttt{ME}}(\ell_2, \varepsilon, b)$ achieved by public-coin schemes), is characterized as follows:

(i) if $\log d <  b  < d$, then 
$$ \tilde{r}_{\texttt{ME}}(\ell_2, \varepsilon, b) \preceq \frac{d}{n \min\lp \varepsilon^2, \varepsilon, b/\log d, d \rp};$$

(ii) if $ b \geq d$, then
$$ \tilde{r}_{\texttt{ME}}(\ell_2, \varepsilon, b) \preceq \frac{d}{n \min\lp \varepsilon^2, \varepsilon, d \rp},$$

and the above errors can be achieved by private-coin SQKR. Therefore private-coin SQKR requires $O\lp \min\lp \lceil \varepsilon \rceil \log d, d \rp \rp$ bits of communication to achieve $\tilde{r}_{\texttt{ME}}\lp \ell_2, \varepsilon, \infty\rp$.
\end{corollary}
Similarly, the estimation error of private-coin RHR is characterized below: 
\begin{corollary}[Private-coin RHR]
Under $\varepsilon$-LDP and $b$-bit communication constraints with $b > \log d$ and $0<\varepsilon \leq \log d$, the $\ell_2$ minimax error for private-coin frequency estimation, denoted as $\tilde{r}_{\texttt{FE}}(\ell_2, \varepsilon, b)$, is
$$ \tilde{r}_{\texttt{FE}}(\ell_2, \varepsilon, b) \preceq\frac{d}{n \min\lp \lp e^\varepsilon-1\rp^2, e^\varepsilon, d \rp},$$
which can be achieved by private-coin RHR. In words, for any $\varepsilon$, private-coin RHR always uses $\log d$ bits of communication to achieve $\tilde{r}_{\texttt{FE}}(\ell_2, \varepsilon, \infty)$.
\end{corollary}

Moreover, the following lemma,  an extension of \cite[Theorem~4]{acharya2019communication}, establishes a lower bound on the communication required for consistent private-coin schemes:
\begin{lemma}\label{lemma:impossibility2}
    Any consistent private-coin scheme for both mean estimation and frequency estimation uses at least $b > \log d -2$ bits of communication.
\end{lemma}
This shows that the $\log d$ lower bounds on $b$ in both corollaries are fundamental (within $2$ bits). The proof of the lemma is given in Section~\ref{sec:impossibility}.
}

%% file: sec_experiments.tex
In this section, we implement our mean estimation and frequency estimation schemes and present our experimental results\footnote{The code can be found in \url{https://github.com/WeiNingChen/Kashin-mean-estimation} (for the SQKR scheme) and \url{https://github.com/WeiNingChen/RHR} (for the RHR scheme).}. More detailed results can be found in Section~\ref{sec:exp_app}.

\subsection{Mean estimation}
We implement our mean estimation scheme Subsampled and Quantized Kashin's Response (SQKR) as in Section~\ref{sec:mean_est} under \emph{private-coin setting} and compare it with a baseline, a concatenation of \texttt{DJW} \cite{duchi2013local, bhowmick2018protection} (which is order-optimal under $\varepsilon$-LDP for $\varepsilon = O(1)$) and the quantizer based on Kashin's representation \cite{lyubarskii2010uncertainty} (which is optimal up to a logarithmic factor, under $b$-bit communication constraint). 

\texttt{DJW} (Lemma~1 in \cite{duchi2013local}) samples a vector from the unit sphere with proper probability density (which depends on $X_i$), and scales it by a factor of $O(\sqrt{d})$ in order to make it unbiased. Although under public-coin setting, one can sample the vector with the help of public randomness and reduce the communication to $\lceil \varepsilon \rceil$ bits \cite{Bassily2017}, for private-coin model each client has to send a $d$-dimensional vector to the server and hence requires to communicate $\Theta(d)$ bits\footnote{We remark that after our paper being published, a recent work \cite{feldman2021lossless} shows that \texttt{DJW} and its improved version \texttt{privUnit} \cite{bhowmick2018protection} can be compressed in a more efficient way. We refer the reader to \cite{feldman2021lossless} for more details.}. To compare with SQKR under private-coin setting, we use an (order-optimal) quantizer based on Kashin's representation to further compress the communication   to $b\lceil \log d\rceil$ bits. It can be shown that such direct concatenation will result in $\tilde{O}(d^2)$ error rate (see Section~\ref{sec:exp_app} in appendix for more details).

\begin{figure}[htbp]
  	\centering
  	\subfloat{{\includegraphics[width=0.49\linewidth]{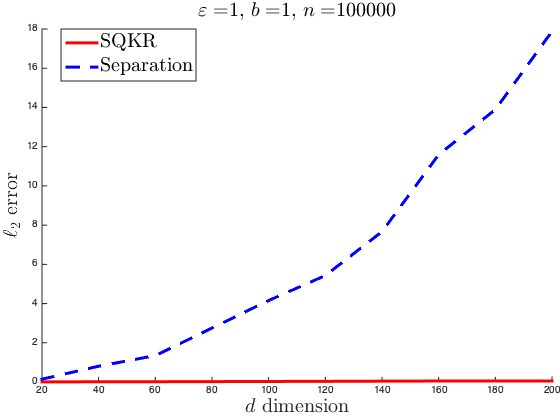} }}%
  	\subfloat{{\includegraphics[width=0.49\linewidth]{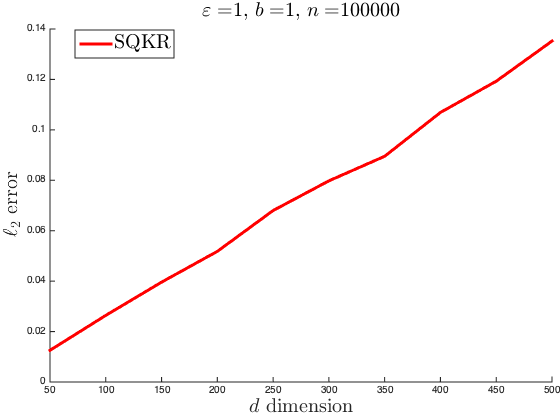} }}%
  	\caption{ $\ell_2$ error with $n=10^5$ and different dimensions $d$. Note that under private-coin setting, the communication cost is $b\lceil \log d \rceil$ bits for both schemes. In order to better emphasize the dependence to $d$, on the right-hand side we only plot the $\ell_2$ error of SQKR.}%
  	\label{fig:mean_est_1}
\end{figure}

\paragraph{Generating the data} In order to capture the distribution-free setting, we generate data independently but non-identically; in particular, we set $Z_1,...,Z_{n/2} \diid N(1,1)^{\otimes d}$ and $Z_{n/2+1},...,Z_{n} \diid N(10,1)^{\otimes d}$ (this also makes the data non-central, i.e. $\E\lb \sum Z_i \rb \neq 0$). Since each sample has bounded $\ell_2$ norm, we normalize each $Z_i$ by setting $X_i = Z_i/\lV Z_i \rV_2$.

\paragraph{Generating the tight frame}
We construct the tight frame by using the random partial Fourier matrices in \cite{lyubarskii2010uncertainty}. Specifically, we set $N = 2^{\lceil \log_2 d \rceil+1} = \Theta(d)$, and choose the basis $U = \lbp 1/\sqrt{N}, -1/\sqrt{N} \rbp^{N \times d}$ by selecting the first $d$ rows of $H_N\cdot D$, where $H_N$ is a $N\times N$ Hadamard matrix and $D$ is a random diagonal matrix with each diagonal entry generated from $\msf{uniform}\lbp  +1, -1\rbp$. It can be shown that the tight frame based on $U$ has Kashin's level $K = \tilde{O}(1)$. 

In Figure~\ref{fig:mean_est_1}, we fix the sample size to $n = 10^5$ and $\varepsilon, b$, and increase the dimension $d$. From the result, we see that SQKR has linear dependence on $d$, whereas the baseline (labeled as "Separation" since it is based on the idea of separately coding for privacy and communication efficiency) has super-linear dependence. Therefore the performance differs drastically when $d$ increases.

\subsection{Frequency estimation}

\begin{figure}[htbp]
  	\centering
  	\subfloat{{\includegraphics[width=0.49\linewidth]{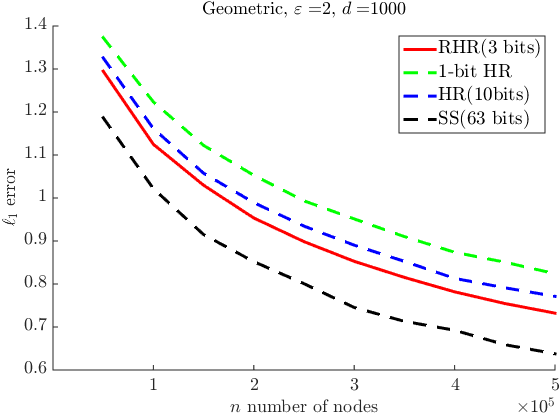} }}%
  	\subfloat{{\includegraphics[width=0.49\linewidth]{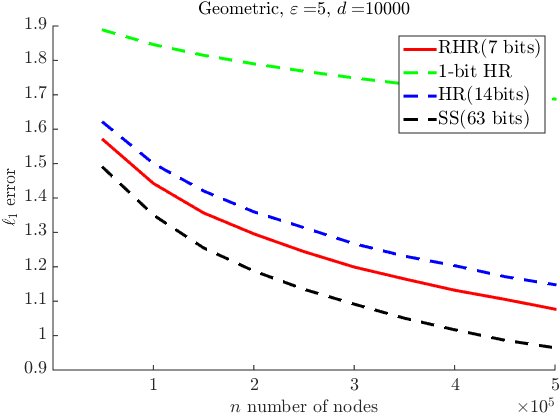} }}%
  	\caption{ $\ell_1$ error with $d = 5000$ and $d=10000$, under (truncated) $Geo(0.8)$ and different $\varepsilon$.}%
  	\label{fig:freq_est_1}
\end{figure}

For frequency estimation problem, we experimentally compare our scheme, Recursive Hadamard Response (RHR), with SS \cite{ye2017optimal}, HR \cite{acharya2019hadamard} and $1$-bit HR \cite{acharya2019communication}\footnote{For HR, we use the codes from \cite{acharya2019hadamard} (\url{https://github.com/zitengsun/hadamard_response})}. We set $d = \{1000, 10000\}$, $\varepsilon = \{ 2, 5\}$, and evaluate the $\ell_1$ estimation errors on the truncated and normalized geometric distribution with $\lambda = 0.8$. For each point (i.e. for each parameter $n, \varepsilon, d$), we repeat the simulation $30$ times and average the $\ell_2$ errors. Figure~\ref{fig:freq_est_1} shows that our schemes can achieve the same performance as HR but is significantly more communication efficient. For instance, in Figure~\ref{fig:freq_est_1} with $d=10000, \varepsilon = 5$, RHR uses only half of the communication budget for HR and achieves better performance.  In all settings, SS has the best statistical performance, but this comes with drastically higher communication and computation cost.

%% file: sec_conclusion.tex
We have investigated mean estimation and frequency estimation under $\varepsilon$-LDP and $b$-bit communication constraints. A significant advantage of the approaches we presented is that they achieve the privacy and communication constraints simultaneously at the cost of the harsher one. Many interesting questions remain to be addressed, including investigating if we can reduce the amount of shared randomness, deriving decoding schemes with optimal runtimes, and applying our results to distributed SGD.

%% file: appendix.tex
\section{Separate Quantization and Privatization Is Strictly Sub-optimal} \label{sec:separation_is_suboptimal}

\paragraph{Distribution estimation}
First let us recap the subset selection (SS) scheme proposed by \cite{ye2017optimal}. Assume $X_1,...,X_n \diid \bm{p} = (p_1,...,p_d)$. Client $i$ maps the local data $X_i$ into $y\in\mcal{Y}_{d, w}\eqDef \lbp y\in \lbp0,1\rbp^d: \sum_j y_j = w\rbp$ with the transitional probability
\begin{equation*}
     Q_{\text{SS}}(y|X=j) = \frac{e^\varepsilon y_j + (1-y_j)}{e^\varepsilon {d-1 \choose w-1}+{d-1 \choose w}}. 
\end{equation*}

The estimator for $p_j$ is defined by 
\begin{equation}\label{eq:SS_estimator}
    \hat{p}_j \eqDef \lp \frac{(d-1)e^\varepsilon +\frac{(d-1)(d-w)}{w}}{(d-w)(e^\varepsilon-1)}\rp \frac{T_j}{n}-\frac{(w-1)e^\varepsilon+d-w}{(d-w)e^\varepsilon -1},
\end{equation} 
where $T_j \eqDef \sum_{i=1}^n Y_i(j)$. Note that by picking $w = \lceil\frac{d}{e^{\varepsilon}+1}\rceil$, SS is order-optimal for all privacy regimes. 

To demonstrate that separating privatization and quantization is strictly sub-optimal, we analyze the estimation error of directly concatenating the $2^b$-SS mechanism with the grouping-based quantization in \cite{han2018distributed}. Note that both schemes are known to be optimal under the corresponding constraints, privacy and communication respectively. However, their direct combination yields an $\ell_2$ error of order $O\lp d^2 \rp$, which is far from the optimal accuracy established in Theorem~\ref{thm:emp_dist_estimation}.

We first group $[d]$ into $s = d/2^b$ equal-sized groups $\mcal{G}_1,...,\mcal{G}_s$, and each client is only responsible to send information about one particular group. That is, let $Y_i$ be the outcome of the $2^b$-SS mechanism, i.e. $Y_i \sim Q_{\text{SS}}\lp \cdot |X_i\rp$, and client $i$ only transmits $\lbp Y_i(j) | j \in \mcal{G}_{s'} \rbp$, for some $s' \in [s]$. Since the server estimates each component of $\bm{p}$ separately as in \eqref{eq:SS_estimator}, this grouping strategy reduces the effective sample size from $n$ to $n' = n 2^b/d$. Plugging $n'$ into the $\ell_2$ error (see Proposition~III.1 in \cite{ye2017optimal}), we conclude that the error grows as 
$$ O\lp\frac{d^2}{n2^b \min\lp e^\epsilon, (e^\epsilon-1)^2 \rp}\rp.$$

Note that since each $Y_i$ contains exactly $w$ ones, the required communication budget to describe $\lbp Y_i(j), j\in\mcal{G}_l\rbp$ may be larger than $b$ bits. But this is fine since it implies that even given more than $b$ bits, the estimation error still grows with $d^2$. In Theorem~\ref{thm:dist_estimation}, on the other hand, we show that the optimal $\ell_2$ error  is linear in $d$, so this demonstrates that separate quantization and privatization is sub-optimal.

\paragraph{Mean estimation}
For the mean estimation problem, a straightforward combination is using the \texttt{PrivUnit} mechanism (see Algorithm~1 in \cite{bhowmick2018protection}) to perturb the local data $X_i \in \mcal{B}_d(\bm{0},1)$, and then using \texttt{RandomSampling} quantization in (Theorem~6 in \cite{g2019vqsgd}) to compress the perturbed data. Both schemes are known to be optimal under the corresponding constraints, privacy and communication respectively. (Note that in Section~\ref{sec:exp} we replaced the  \texttt{RandomSampling} quantization with a Kashin's quantizer, since implementing the theoretically optimal \texttt{RandomSampling} quantizaton is computationally infeasible.)

By Proposition~4 in \cite{bhowmick2018protection}, the output of \texttt{PrivUnit}, denoted as $Z_i = \texttt{PrivUnit}\lp X_i, \varepsilon\rp$, has $\ell_2$ norm of order $\Theta\lp \sqrt{\frac{d}{\min\lp \varepsilon, \varepsilon^2 \rp}} \rp$. However, if we further apply \texttt{RandomSampling} to $b$ bits, by Theorem~6 in \cite{g2019vqsgd}, the $\ell_2$ estimation error grows as 
$$ \Theta\lp \left\| Z_i \right\| \frac{d}{n\cdot b} \rp = \Theta\lp \frac{d^2}{n b\min\lp \varepsilon, \varepsilon^2 \rp} \rp, $$
showing a quadratic dependence in $d$. By Theorem~\ref{thm:mean_estimation}, nevertheless, we can construct a better scheme with $O(d/n\min\lp \varepsilon, \varepsilon^2, b \rp)$ dependence under both constraints.

\section{More Experimental Results}\label{sec:exp_app}

\input{sec_exp_app}

\section{Proof of Theorem~\ref{thm:mean_estimation}}\label{sec:mean_est_app}
\input{sec_mean_estimation_app}

\section{Proof of Theorem~\ref{thm:emp_dist_estimation}}\label{sec:freq_est_app}
\input{sec_freq_estimation_app}

\newpage
{
\section{Proofs for Section~\ref{sec:rand}} \label{sec:impossibility}
We start with proving Lemma~\ref{lemma:impossibility2}. Without access to the public randomness, \cite{acharya2019communication} shows that at least $\Theta(d)$ bits of communication is required for heavy hitter estimation in order to obtain a consistent estimator\footnote{Recall that an estimator is consistent if it has vanishing estimation error as $n$ tends to infinity.}. We state their result here:
\begin{lemma}[\cite{acharya2019communication} Theorem~4]\label{lemma:impossibility}
    Let $b \leq \log d -2$. For all private-coin schemes $\lp Q^n, \hat{D}\rp$ with only private randomness and $b$ bits communication budgets, there exists a data sets $X_1,...,X_n$ with $n > 12(2^b+1)^2$, such that
    \begin{equation*}
        \E\lb \left\| \hat{D}\lp Q^n \rp - D_{X^n} \right\|_\infty\rb \geq \frac{1}{2^{b+2}+4}.
    \end{equation*}
\end{lemma}

Based on this, we claim that without public coin, each client needs to transmit at least $\Theta(\log d)$ bits in order to construct consistent schemes for frequency estimation or mean estimation.

\subsection{Proof of Lemma~\ref{lemma:impossibility2}}
\paragraph{Frequency estimation} We lower bound $\ell_1$ and $\ell_2$ error by $\ell_\infty$ and apply Lemma~\ref{lemma:impossibility}.
\begin{align*}
    \E\lb \left\| \hat{D}\lp Q^n \rp - D_{X^n} \right\|_1\rb \geq \E\lb \left\| \hat{D}\lp Q^n \rp - D_{X^n} \right\|_\infty\rb \geq \frac{1}{2^{b+2}+4},
\end{align*}
and 
\begin{align}\label{eq:freq_impossibility_lbd}
    \E\lb \left\| \hat{D}\lp Q^n \rp - D_{X^n} \right\|^2_2\rb 
    & \geq \E\lb \left\| \hat{D}\lp Q^n \rp - D_{X^n} \right\|^2_\infty\rb \nonumber\\
    & \geq \lp \E\lb \left\| \hat{D}\lp Q^n \rp - D_{X^n} \right\|_\infty\rb \rp^2 \nonumber\\
    & \geq \lp \frac{1}{2^{b+2}+4}\rp^2.
\end{align}
This implies that it is impossible to construct consistent schemes with less than $\log d-2$ bits per client in the absence of a public randomness. On the other hand, given $\log d$ bits, one can readily achieve the optimal estimation accuracy without any public randomness, for instance, by using Hadamard response \cite{acharya2019hadamard} (see also the discussion in \cite{acharya2019communication}). Therefore, the problem of frequency estimation is somewhat trivialized in the absence of public randomness.

\paragraph{ Mean estimation} Let $X_i\in[d]$ be one-hot encoded, so $X_i \in \mcal{B}_d\lp\bm{0}, 1\rp$. Then \eqref{eq:freq_impossibility_lbd} implies the $\ell_2$ error of mean estimation is at least $1/\lp2^{b+2}+4\rp^2$. Thus with less than $\log d-2$ bits of communication budget, it is also impossible to construct a consistent scheme for mean estimation. \qedwhite

\subsection{Proof of Corollary~\ref{cor:me_rnd} and Corollary~\ref{cor:me_rnd}}\label{sec:proof_cor_rnd}

Notice that since one can always ``simulate'' the public coin by uplink communication (i.e. each client generates its private random bits and send them to the server), any $b$ bits public-coin scheme can be cast into a private coin scheme with additional $b$ bits communication. This implies the above impossibility results (Lemma~\ref{lemma:impossibility2}) also serves a valid lower bound for the amount of public randomness: for any public-coin scheme with $b < \log d -2$ bits communication budgets, we need at least $\log d -b -2$ bits of shared randomness in order to obtain a consistent estimate of the empirical mean or empirical frequency. \qedwhite
}

\section{Proof of Claims}
\subsection{Proof of Claim~\ref{clm:mse_bd}}
\begin{proof}
According to \eqref{eq:mse_bdd}, it suffices to control $\Var\lp \hat{a}_j\rp$. To bound the variance, consider
\begin{align*}
    \Var\lp \hat{a}_j\rp 
    & = \frac{N^2}{k^2}\cdot\lp \frac{e^\varepsilon +2^k - 1}{e^\varepsilon -1} \rp^2 \Var\lp\sum_{m=1}^k \tilde{q}_m\cdot\bbm{1}_{\lbp j=s_m \rbp}\rp\\
    & \leq  \frac{N^2}{k^2}\cdot\lp \frac{e^\varepsilon +2^k - 1}{e^\varepsilon -1} \rp^2
    \E\lb\lp\sum_{m=1}^k \tilde{q}_m\cdot\bbm{1}_{\lbp j=s_m \rbp}\rp^2\rb \\
    & \overset{\text{(a)}}{\leq} \frac{N^2}{k^2}\cdot\lp \frac{e^\varepsilon +2^k - 1}{e^\varepsilon -1} \rp^2 \lp \frac{c}{\sqrt{d}}\rp^2 \E\lb\lp\sum_{m=1}^k \bbm{1}_{\lbp j=s_m \rbp}\rp^2\rb\\
    & \overset{\text{(b)}}{\leq} C\frac{N}{k^2}\cdot\lp \frac{e^\varepsilon +2^k - 1}{e^\varepsilon -1} \rp^2 \lp \frac{k^2}{N^2}+\frac{k}{N} \rp\\
    &=C\lp \frac{e^\varepsilon+2^k-1}{e^\varepsilon-1}\rp^2\lp\frac{1}{N}+\frac{1}{k}\rp,
\end{align*}
where (a) is due to $\lba\tilde{q}_m\rba = \frac{c}{\sqrt{d}}$, and (b) is due to the second moment bound on $\msf{Binomial}(k, 1/N)$ and the fact $N = \Theta(d)$. Therefore by \eqref{eq:mse_bdd},
$$ \E\lb \left\| \hat{X} - X \right\|_2^2\rb \leq C_0\sum_{i=1}^N\Var\lp\hat{a}_i\rp \leq C_1\lp \frac{e^\varepsilon+2^k-1}{e^\varepsilon-1}\rp^2\frac{d}{k},$$
establishing the claim.
\end{proof}

\subsection{Proof of Claim~\ref{clm:y_unbiased}}
\begin{proof}
$\hat{Y}_i$ yields an unbiased estimator since
\begin{align}\label{eq:Y_i_unbias}
    \E\lb \hat{Y}_i\lp \frac{e^\varepsilon + 2^k -1}{e^\varepsilon -1}\tilde{Q}_i, r_i \rp\rb  
    & = \E\lb\E\lb \hat{Y}_i\lp \frac{e^\varepsilon + 2^k -1}{e^\varepsilon -1}\tilde{Q}_i, r_i \rp\Big| r_i\rb\rb \nonumber\\
    & \overset{\text{(a)}}{=} \E\lb \hat{Y}_i\lp \E \lb\frac{e^\varepsilon + 2^k -1}{e^\varepsilon -1}\tilde{Q}_i\Big| r_i\rb, r_i \rp\rb \nonumber\\
    & = \E\lb \hat{Y}_i\lp Q(X_i, r_i), r_i \rp\rb \nonumber\\
    & = \frac{1}{d}H_d X_i,
\end{align}
where (a) holds since conditioning on $r_i$, $\hat{Y}_i(Q, r_i)$ is a linear function of $Q$.
\end{proof}

\subsection{Proof of Claim~\ref{clm:bdd_err_emp}}
\begin{proof}
The $\ell_2$ error is
\begin{align} \label{eq:bdd_ell_2}
    \E\lb \left\| \hat{D} - D_{X^n} \right\|^2_2 \rb
    & = \frac{1}{n^2}\sum_{i=1}^n \E\lb \left\| H_d \hat{Y}_i - H_d \E\lb \hat{Y}_i \rb \right\|^2_2 \rb \nonumber \\
    & = \frac{d}{n^2}\sum_{i=1}^n \E\lb \left\| \hat{Y}_i - \E\lb \hat{Y}_i \rb \right\|_2^2 \rb. 
\end{align}
It remains to bound $\E\lb \left\| \hat{Y}_i - \E\lb Y_i \rb \right\|_2^2 \rb$. Observe that 
$$\lba \E[\hat{Y}_i]\rba = \lba \frac{H_d\cdot X_i}{d} \rba = [1/d,...,1/d]^\intercal,$$
and from expression \eqref{eq:hat_Y_i}, given $r_i$, there are only $2^{k-1}$ non-zero coordinates, each with value bounded by $ \lp \frac{e^\varepsilon + 2^k -1}{e^\varepsilon -1} \rp / 2^{k-1}$. Therefore we have
\begin{align*}
    \E\lb \left\| \hat{Y}_i - \E\lb \hat{Y}_i \rb \right\|_2^2 \rb 
    & = \E \lb \E\lb \left\| \hat{Y}_i - \E\lb \hat{Y}_i \rb \right\|_2^2 \Big| r_i \rb\rb \\
    & \leq 2\lp d\lp \frac{1}{d} \rp^2 + 2^{k-1} \lp\frac{e^\varepsilon + 2^k -1}{2^{k-1}\lp e^\varepsilon -1\rp}\rp^2\rp.
\end{align*}
Plugging this in to \eqref{eq:bdd_ell_2}, we arrive at
$$ \E\lb \left\| \hat{D} - D_{X^n} \right\|^2_2 \rb \preceq \frac{d}{n 2^{k-1}} \lp\frac{e^\varepsilon + 2^k -1}{\lp e^\varepsilon -1\rp}\rp^2. $$

Picking $k = \min \lp b, \lceil \varepsilon \log_2 e \rceil, \lfloor \log d \rfloor\rp$ yields 
\begin{align*}
    \E\lb \left\| \hat{D} - D_{X^n} \right\|^2_2\rb
    & = O\lp \frac{d}{n \min\lp 2^b, e^\varepsilon, d\rp} \lp \frac{e^\varepsilon}{e^\varepsilon - 1} \rp^2 \rp.
\end{align*}
Observe that 
\begin{itemize}
    \item[(i)] if $e^\varepsilon = O(2^b)$, then $e^\varepsilon \preceq 2^b$, so 
    $\E\lb \left\| \hat{D} - D_{X^n} \right\|^2_2\rb = O\lp \frac{d e^\varepsilon}{n\lp e^\varepsilon - 1\rp^2} \rp.$
    \item[(ii)] If $e^\varepsilon = \Omega(2^b)$, then $\frac{e^\varepsilon}{e^\varepsilon -1} = \theta(1)$, and
    $\E\lb \left\| \hat{D} - D_{X^n} \right\|^2_2\rb = O\lp \frac{d}{n \min \lp 2^b, d \rp} \rp.$
\end{itemize}

Therefore we conclude that 
$$\E\lb \left\| \hat{D} - D_{X^n} \right\|^2_2\rb \preceq \max \lp \frac{d}{n \min\lp 2^b, d\rp}, \frac{d e^\varepsilon}{n\lp e^\varepsilon -1\rp^2} \rp \asymp \frac{d}{n} \lp \frac{1}{\min{\lbp e^\varepsilon, \lp e^{\varepsilon} -1 \rp^2, 2^b, d\rbp}} \rp.$$

By Jensen's inequality and Cauchy-Schwarz inequality, we also have
\begin{align*}
    \E \lb \left\| \hat{D} - D_{X^n} \right\|_1 \rb
    &\leq \lp \E\lb\left\| \hat{D} - D_{X^n} \right\|_1^2\rb\rp^{\frac{1}{2}}
    \leq \lp d\cdot\E\left\| \hat{D} - D_{X^n} \right\|_2^2\rp^{\frac{1}{2}}\\
    &\preceq \frac{d}{\sqrt{n\lp \min{\lbp e^\varepsilon, \lp e^{\varepsilon} -1 \rp^2, 2^b, d\rbp} \rp}}. 
\end{align*}
\end{proof}

%% file: sec_exp_app.tex
\subsection{Mean estimation}
We generate the data as well as the tight frame as described in Section~\ref{sec:exp}.
\paragraph{Compare to optimal LDP estimation schemes}
We first compare our scheme SQKR, under private-coin setting, with 1) \texttt{privUnit} \cite{bhowmick2018protection}, which is order-optimal for all $\varepsilon$ and 2) \texttt{DJW} \cite{duchi2013local}, which is order-optimal for $\varepsilon = O(1)$. Note that although \texttt{DJW} is originally designed for high-privacy regime $\varepsilon = O(1)$, one can independently and repeatedly apply it with $\varepsilon' = 1$ for $\lfloor \varepsilon \rfloor$ times and return the mean of the $\lfloor \varepsilon \rfloor$ vectors. By the composition theorem \cite{kairouz2015composition}, the output satisfies $\lfloor \varepsilon \rfloor$-LDP, and the MSE is reduced by a factor of $\lfloor \varepsilon \rfloor$. The repeated version of \texttt{DJW} (denoted as \texttt{reDJW}) is hence asymptotically optimal, and we also compare it with our scheme.  

\begin{figure}[htbp]
  	\centering
  	\subfloat{{\includegraphics[width=0.49\linewidth]{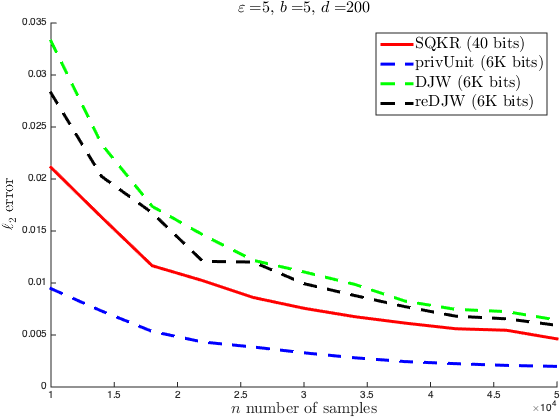} }}%
  	\subfloat{{\includegraphics[width=0.49\linewidth]{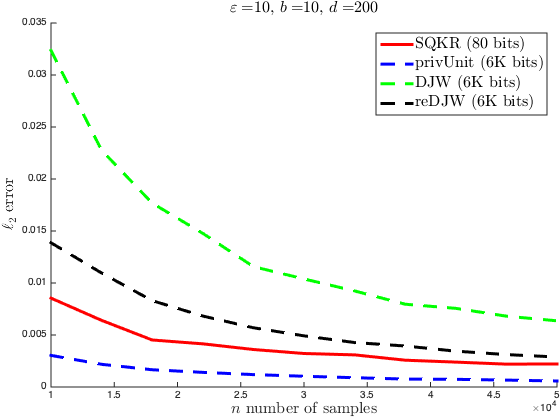} }}%
  	\caption{ $\ell_2$ error of \texttt{privUnit}, \texttt{DJW}, \texttt{reDJW} and SQKR with different dimensions $d= 200$.}
  	\label{fig:priv}
\end{figure}

Note that the outcomes of \texttt{privUnit}, \texttt{DJW} and \texttt{reDJW} are $d$-dimensional vectors lying in a radius $O(\sqrt{d})$ sphere, so in general we need $32d$ bits to represent it (where we assume each float requires $32$ bits). Figure~\ref{fig:priv} shows that SQKR achieves similar performance with significantly communication budgets. For instance, under private-coin model, when $\varepsilon = 5$ and $d = 200$, the communication cost of \texttt{privUnit} is roughly $32 \times 200 \approx 6K$ bits, while according to Corollary~\ref{cor:priv_SQKR}, SQKR uses only $5\times \lceil\log_2 200\rceil = 40 $ bits.

Next, under private-coin setting, we compare SQKR with a combination of \texttt{DJW} and an optimal quantizer.
\paragraph{Baseline: a direct concatenation of \texttt{DJW}, Kashin's quantizer and sampling}
For each $X_i$ in unit $\ell_2$ ball, \texttt{DJW} maps it to a vector $\tilde{X}_i$ with length $\lV \tilde{X}_i \rV_2 = \Theta\lp \sqrt{d/\min\lp 1, \varepsilon^2 \rp} \rp$. Note that $\texttt{DJW}$ is order-optimal for $\varepsilon = O(1)$ \cite{duchi2013local}. If we quantize $\tilde{X}_i$ according to its Kashin's representation and then subsample $b$ bits from it as in Section~\ref{sec:mean_est}, then the $\ell_2$ error (i.e. variance) will be 
$$ \tilde{O}\lp \frac{d}{b} \lV \tilde{X}_i \rV^2 \rp =  \tilde{O}\lp \frac{d^2}{b \min\lp 1, \varepsilon^2 \rp}\rp.$$
Therefore, averaging over $n$ clients, the $\ell_2$ error of estimating the empirical mean is
$$ \tilde{O}\lp \frac{d^2}{ n \cdot b \min\lp 1, \varepsilon^2 \rp}\rp.$$

However, in Theorem~\ref{thm:mean_estimation}, we see that with a more sophisticated design, we can achieve smaller $\ell_2$ error 
$$ O\lp \frac{d}{ n \cdot \min\lp \varepsilon, \varepsilon^2, b \rp}\rp.$$

\paragraph{Setup}
In the experiment, we mainly focus on the \emph{high-privacy low-communication} setting where $\varepsilon = b = 1$. Note that since we are under private-coin setting, the actual communication cost for each setting is $b\cdot \lceil \log_2 d \rceil$. 

\begin{figure}[htbp]
  	\centering
  	\subfloat{{\includegraphics[width=0.43\linewidth]{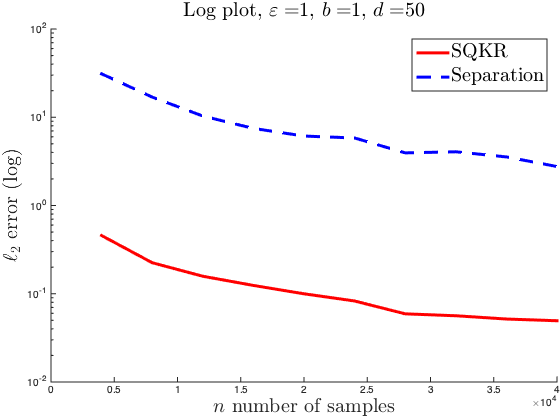} }}%
  	\subfloat{{\includegraphics[width=0.43\linewidth]{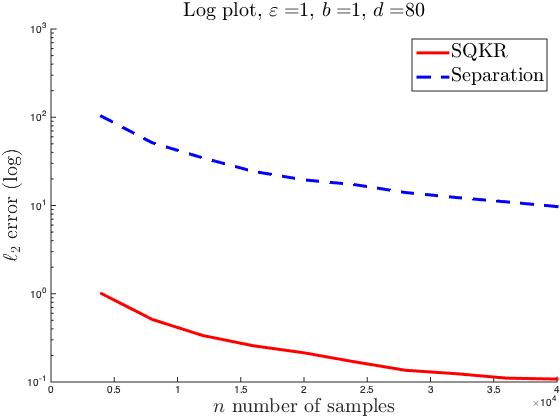} }}%
  	\caption{ Log-scale $\ell_2$ error with different dimensions $d= 50, 80$ and different privacy and communication budgets.}
  	\label{fig:mean_est_2}
\end{figure}

We first consider different dimensions $d$ and plot the (log-scale) $\ell_2$ estimation error (i.e. mean square error) with sample size $n$. For each point, i.e. each set of parameters $(\varepsilon, b, d, n)$, we repeat the simulation for $8$ iterations and report the average. In Figure~\ref{fig:mean_est_2}, we see that SQKR drastically outperforms the baseline (labeled as "Separation" since it is based on the idea of separately coding for privacy and communication efficiency). The gain increases in higher dimensions or with more stringent privacy/communication constraints.

\begin{figure}[htbp]
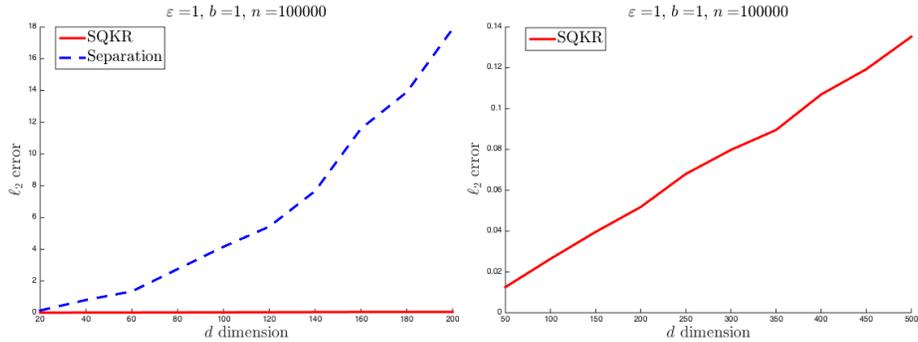

  	\centering
  	\subfloat{{\includegraphics[width=0.43\linewidth]{./mean_est_data/plot_dimension_eps_1_d_200} }}%
  	\subfloat{{\includegraphics[width=0.43\linewidth]{./mean_est_data/plot_dimension_eps_1_d_200_kashin} }}%
  	\caption{ $\ell_2$ error with $n=10^5$ and different dimensions $d$. In order to better emphasize the dependence to $d$, on the right-hand side we only plot the $\ell_2$ error of SQKR.}%
  	\label{fig:mean_est_3}
\end{figure}

Next, to better study the dependence on $d$, we fix the sample size to $n = 10^5$ and $\varepsilon= b=1$, and increase the dimension $d$. In Figure~\ref{fig:mean_est_3}, We see that SQKR has linear dependence on $d$, and \texttt{Separation} has super-linear dependence. Therefore the performance differs drastically when $d$ increases.
\newpage 
\subsection{Frequency estimation}

\begin{figure}[htbp]
  	\centering
  	\subfloat{{\includegraphics[width=0.49\linewidth]{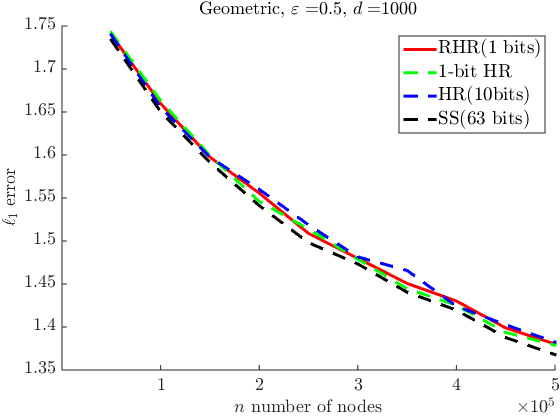} }}%
  	\subfloat{{\includegraphics[width=0.49\linewidth]{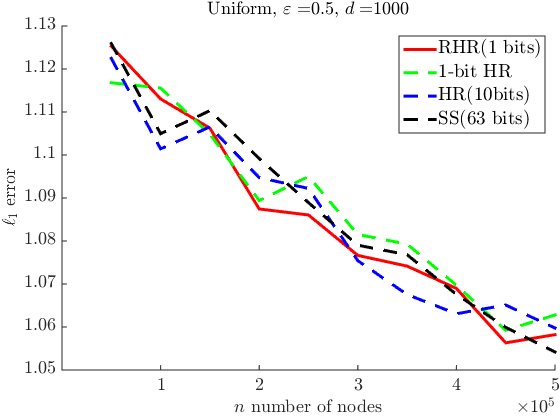} }}%
  	\newline
  	\subfloat{{\includegraphics[width=0.49\linewidth]{./data_clip_v1/data_clip_Geometric_k_1000_eps_2_.png} }}%
  	\subfloat{{\includegraphics[width=0.49\linewidth]{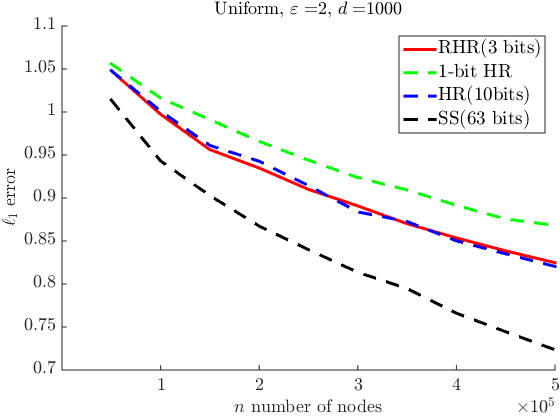} }}%
  	    \newline
  	\subfloat{{\includegraphics[width=0.49\linewidth]{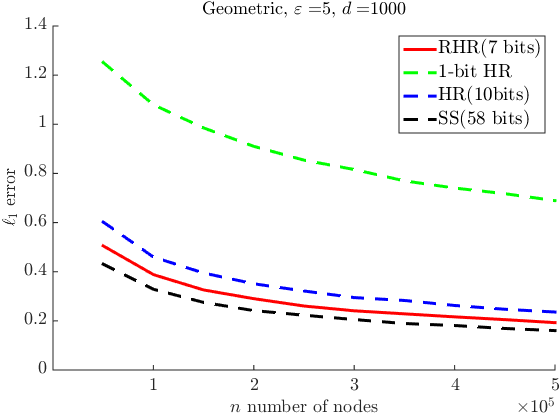} }}%
  	\subfloat{{\includegraphics[width=0.49\linewidth]{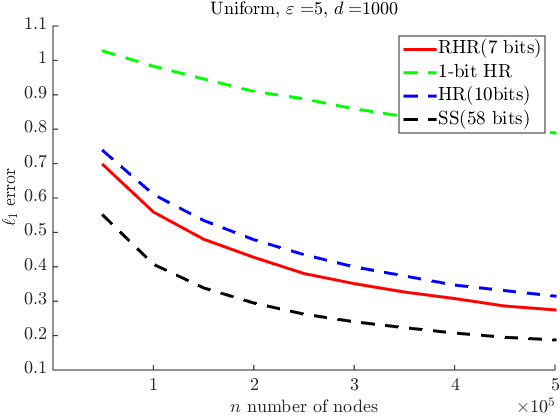} }}%
  	\caption{ $\ell_1$ error with $d = 1000$. Left are $Geo(0.8)$ and right are \textit{Uniform}.}%
  	\label{fig:dist_emp_1000}
\end{figure}

\begin{figure}[htbp]
  	\centering
  	\subfloat{{\includegraphics[width=0.49\linewidth]{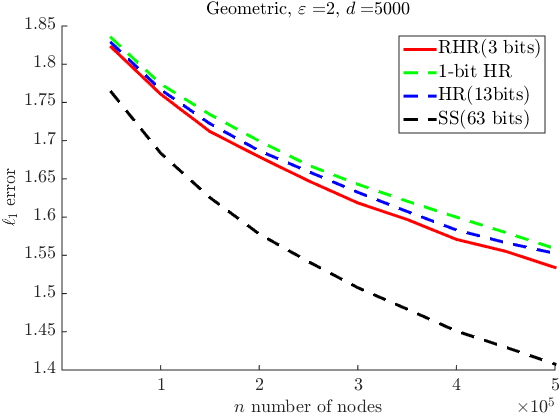} }}%
  	\subfloat{{\includegraphics[width=0.49\linewidth]{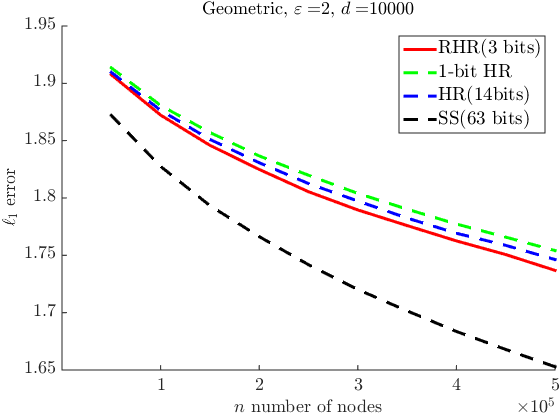} }}%
  	    \newline
  	\subfloat{{\includegraphics[width=0.49\linewidth]{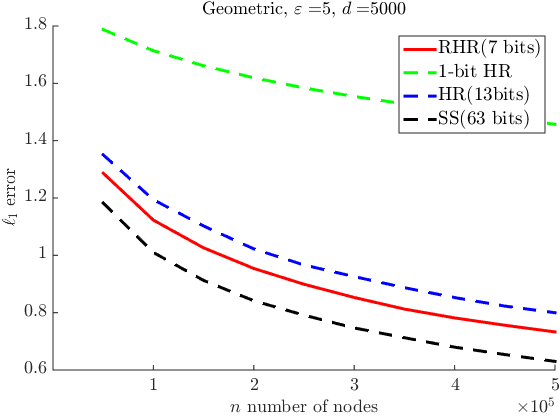} }}%
  	\subfloat{{\includegraphics[width=0.49\linewidth]{./data_clip_v1/data_clip_Geometric_k_10000_eps_5_2.png} }}%
  	\caption{ $\ell_1$ error with $d = 5000$ and $d=10000$, under (truncated) $Geo(0.8)$ and different $\varepsilon$.}%
  	\label{fig:dist_emp_5000}
\end{figure}

For frequency estimation, we compare our scheme, Recursive Hadamard Response (RHR), with SS \cite{ye2017optimal}, HR \cite{acharya2019hadamard} and $1$-bit HR \cite{acharya2019communication}. We set $d = \lbp 1000, 5000, 10000 \rbp$, $\varepsilon \in \lbp 0.5, 2, 5 \rbp$ and $n = \lbp 50000, 100000, ..., 500000 \rbp$, and evaluate the $\ell_1$ estimation errors on uniform distribution and truncated and normalized geometric distribution with $\lambda = 0.8$. For each point (i.e. for each parameter $n, \varepsilon, d$), we repeat the simulation $30$ times and average the $\ell_2$ errors. Figure~\ref{fig:dist_emp_1000} and Figure~\ref{fig:dist_emp_5000} show that RHR can achieve the same performance as HR but is significantly more communication efficient. For instance, in Figure~\ref{fig:dist_emp_5000} with $d=10000, \varepsilon = 5$, RHR uses only half of the communication budget for HR and achieves better performance.  In all settings, $k$-SS has the best statistical performance, but this comes with drastically higher communication and computation cost.


\begin{figure}[htbp]
  	\centering
  	\subfloat{{\includegraphics[width=0.49\linewidth]{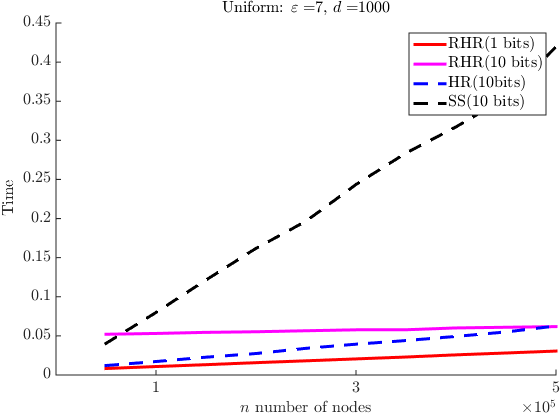} }}%
  	\subfloat{{\includegraphics[width=0.49\linewidth]{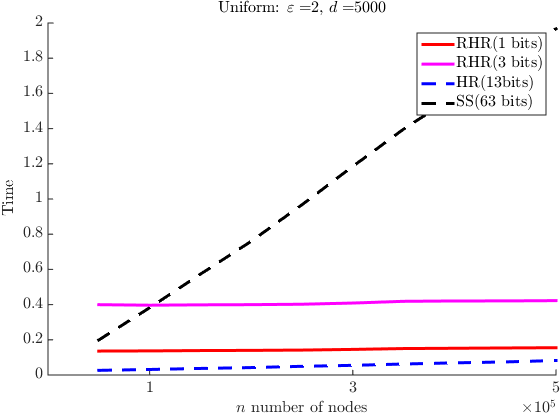} }}%
  	\caption{Left: time complexity with $d = 1000, \varepsilon = 7$  right: time complexity with $d = 5000, \varepsilon = 2$.}%
  	\label{fig:freq_est_computation}
\end{figure}

\newpage
In Figure~\ref{fig:freq_est_computation}, we record the decoding time for each scheme. The decoding complexity of RHR is similar to HR and $1$-bit HR, which are all much more computationally efficient than SS.

\newpage

%% file: sec_mean_estimation_app.tex
\subsection{Achievability}
In this section, we prove that Subsampled and Quantized Kashin's Response (SQKR) achieves optimal $\ell_2$ estimation error.
For each observation $X_i$, we will construct an unbiased estimator $\hat{X}_i$ (i.e. $\E\lb \hat{X}_i | X_i\rb = X_i$), where $\hat{X}_i$ is $\varepsilon$-LDP, can be described by $k$ bits, and has small variance. The encoding scheme consists of three main steps: (1) obtaining a Kashin's representation for a tight frame \cite{lyubarskii2010uncertainty}, (2) subsampling and (3) privatization.

\paragraph{ Kashin's representation}
We begin with introducing tight frames and Kashin's representation \cite{lyubarskii2010uncertainty}.
\begin{definition}[Tight frame]
A tight frame is a set of vectors $\lbp u_j\rbp^N_{j=1} \in \mbb{R}^d$ that obeys Parseval's identity
$$ \left\| x \right\|^2_2 = \sum_{j=1}^N \lan u_j, x \ran^2, \, \text{ for all } x \in \mbb{R}^d. $$
\end{definition}
A frame can be viewed as a generalization of an orthogonal basis in $\mbb{R}^d$, which can improve the encoding stability by adding redundancy to the representation system when $N>d$. To increase robustness, we wish the information to spread evenly in each coefficient, which motivates the following  definition of  a Kashin's representation:
\begin{definition}[ Kashin's representation]
    For a set of vectors $\lbp u_j\rbp_{j=1}^N$, we say the expansion 
    $$ x = \sum_{j=1}^N a_ju_j, \text{ with } \max_j \lba a_j \rba \leq \frac{K}{\sqrt{N}}\lV x \rV_2 $$
    is a Kashin's representation of vector $x$ at level $K$ . 
\end{definition}
Therefore, if we can obtain unbiased estimators $\lbp \hat{a}_j \rbp_{j=1}^N$ of the Kashin's representation of $X $ with respect to a tight frame $\lbp u_j \rbp_{j=1}^N$, then the MSE can be controlled by
\begin{align}\label{eq:mse_bdd}
    \E\lb \lp \hat{X}  - X  \rp^2 \rb 
      = \E\lb \lV \sum_{j=1}^N \lp \hat{a}_j-a_j\rp u_j \rV_2^2 \rb 
     \overset{\text{(a)}}{\leq}  \E\lb  \sum_{j=1}^N \lp \hat{a}_j-a_j\rp^2 \rb 
      =  \sum_{j=1}^N \Var\lp \hat{a}_j\rp,
\end{align}
where (a) is due to the Cauchy–Schwarz inequality and the definition of a tight frame. Recall that $X$ is deterministic, so here the expectation is taken with respect to the randomness on $\hat{a}_j$. Notice that the cardinality $N$ of the frame determines the compression (i.e. quantization) rate, and Kashin's level $K$ affects the variance. Hence we are interested in constructing tight frames with small $N$ and $K$.

By Theorem~3.5 and Theorem~4.1 in \cite{lyubarskii2010uncertainty}, we have the following lemma:
\begin{lemma}[Uncertainty principle and  Kashin's Representation]\label{lemma:up_ Kashin}
For any $\mu>0$ and $N > (1+\mu)d$, there exists a tight frame $\lbp u_j \rbp_{j=1}^N$ with  Kashin's level $K = O\lp \frac{1}{\mu^3}\log\frac{1}{\mu}\rp$. Moreover, for each $X $, finding  Kashin's coefficient requires $O\lp dN\log N \rp$ computation.
\end{lemma}
For our purpose, we choose $\mu$ to be a constant, i.e. $\mu = \Theta(1)$, so $N = \Theta(d), K = \Theta(1)$, and we can obtaina representation of $X  = \sum_{j=1}^N a_j u_j$, with $\lba a_j \rba \leq \frac{K}{\sqrt{N}} = \frac{c}{\sqrt{d}}$ for some constant $c$. Therefore, we quantize each  $a_j$ as follows:
\begin{equation}
      q_j \eqDef 
    \begin{cases}
    -\frac{c}{\sqrt{d}}, \text{ with probability } \frac{c/\sqrt{d} - a_j}{2c/\sqrt{d}}\\
    \frac{c}{\sqrt{d}}, \text{ with probability } \frac{a_j + c/\sqrt{d}}{2c/\sqrt{d}}.
    \end{cases}
\end{equation}
$\bm{q} \eqDef \lp q_1,...,q_N \rp$ yields an unbiased estimator of $\bm{a} \eqDef (a_1,...,a_N)$ and can be described by $N = \Theta(d)$ bits.

\paragraph{Sampling}
To further reduce the communication cost, we sample $k$ bits uniformly at random from $ \bm{q}$ using public randomness. Let $s_1,...,s_k \diid \text{uniform}[N]$ be the indices of the sampled elements, and define the sampled message as
\begin{equation}
    Q \lp \bm{q}, \lp s_1,...,s_k \rp\rp = \lp q_{s_1},...,q_{s_k}\rp \in \lbp -c/\sqrt{d}, c/\sqrt{d} \rbp^k.
\end{equation}
Then $Q $ can be described in $k$ bits, and each of $q_{s_m}$ yields an independent and unbiased estimator of $\bm{a}$:
\begin{align}\label{eq:unbiased_1}
   \E\lb N\cdot q_{s_m}\cdot\bbm{1}_{\lbp j=s_m\rbp}\rb = \E\lb\E\lb N\cdot q_{s_m}\cdot\bbm{1}_{\lbp j=s_m\rbp}\mv q_1,...,q_N\rb\rb = \E\lb q_j \rb = a_j, \,\forall j\in[N].
\end{align}

\paragraph{Privatization}
Each client then perturbs $Q $ via $2^k$-RR mechanism (as a $k$-bit string):
\begin{equation}\label{eq:me_privatize}
    \tilde{Q}  = 
    \begin{cases}
    Q , \text{ with probability } \frac{e^\varepsilon}{e^\varepsilon + 2^k -1}\\
    Q' \in \lbp -c/\sqrt{d}, c/\sqrt{d} \rbp^k / \lbp Q  \rbp, \text{ with probability }\frac{1}{e^\varepsilon + 2^k -1}.
    \end{cases}
\end{equation}
Since 
$$\sum_{Q'\in \lbp -c/\sqrt{d}, c/\sqrt{d} \rbp^k / \lbp Q  \rbp}Q' = -Q ,$$
it is not hard to see 
$ \lp \frac{e^\varepsilon +2^k - 1}{e^\varepsilon -1}\rp \tilde{Q}  $ yields an unbiased estimator of $Q $. Indeed, if we write $\tilde{Q}  = \lp \tilde{q}_1,...,\tilde{q}_k \rp$, then
\begin{equation}\label{eq:unbiased_2}
    \E\lb \lp\frac{e^\varepsilon +2^k - 1}{e^\varepsilon -1}\rp \cdot \tilde{q}_m \mv q_1,...,q_N, s_1,...,s_k \rb = q_{s_m},
\end{equation}
or equivalently 
$$ \E\lb \lp\frac{e^\varepsilon +2^k - 1}{e^\varepsilon -1}\rp\tilde{Q}  \mv Q  \rb = Q .$$

\paragraph{Estimation and the $\ell_2$ error}
Given $\tilde{Q} = \lp \tilde{q}_1,...,\tilde{q}_k \rp$, define
$$ \hat{a}_j = \frac{N}{k}\cdot\lp \frac{e^\varepsilon +2^k - 1}{e^\varepsilon -1} \rp \sum_{m=1}^k \tilde{q}_m\cdot\bbm{1}_{\lbp j=s_m \rbp}.$$
According to \eqref{eq:unbiased_1} and \eqref{eq:unbiased_2}, $\E\lb \hat{a}_j \rb = a_j$, and hence $\hat{X}\lp \tilde{Q}, \lp s_1,...,s_k\rp\rp \eqDef \sum_{j=1}^N \hat{a}_j u_j$ gives us an unbiased estimator of $X$. 

\begin{claim}\label{clm:mse_bd}
The MSE of $\hat{X}$ can be bounded by
$$ \E\lb \left\| \hat{X} - X \right\|_2^2\rb \leq C\lp \frac{e^\varepsilon+2^k-1}{e^\varepsilon-1}\rp^2\frac{d}{k}.$$
\end{claim}

Finally, each client encodes its data $X_i$ independently, and the server computes $\frac{1}{n}\sum_i \hat{X}_i$. Since $\hat{X}_i$ is unbiased and by Claim~\ref{clm:mse_bd}, we get
$$ \E\lb \left\| \frac{1}{n}\sum_{j=1}^n \hat{X}_i - \bar{X} \right\|^2_2 \rb 
= \frac{1}{n^2}\sum_{j=1}^n \E\lb \left\| \hat{X}_i - X_i \right\|_2^2\rb \leq C\lp \frac{e^\varepsilon+2^k-1}{e^\varepsilon-1}\rp^2\frac{d}{nk}.$$
Finally, picking $k = \min\lp \lceil\log_2 e \rceil\varepsilon, b \rp$ gives us the desired upper bound.

\subsection{Lower Bound of Theorem~\ref{thm:mean_estimation}}
As in the converse part of Theorem~\ref{thm:emp_dist_estimation}, the lower bound can be obtained by constructing a prior distribution on $X_i$ and analyzing the statistical mean estimation problem. Therefore, we will impose a prior distribution $P$ on $X_1,..., X_n$ and lower bound the $\ell_2$ error of estimating  the mean $\theta(P)$, where $P$ is a distribution supported on the $d$-dimension unit ball. 

For any $\hat{X}$, observe that
\begin{align}\label{eq:mean_est_lower}
    \E_{\hat{X}, X^n\diid P}\lb \left\| \hat{X} - \bar{X} \right\|_2^2 \rb 
    & \overset{\text{(a)}}{\geq} \E\lb \lp \left\| \hat{X} - \theta\lp P\rp \right\|_2 - \left\| \bar{X} - \theta\lp P\rp \right\|_2 \rp^2 \rb \nonumber\\
    & \geq \E\lb \left\| \hat{X} - \theta\lp P\rp \right\|^2_2 \rb  -2 \E\lb \left\| \hat{X} - \theta\lp P\rp \right\|_2\left\| \bar{X} - \theta\lp P\rp \right\|_2 \rb \nonumber\\
    & \overset{\text{(b)}}{\geq} \E\lb \left\| \hat{X} - \theta\lp P\rp \right\|^2_2 \rb  -2 \sqrt{ \E\lb \left\| \hat{X} - \theta\lp P\rp \right\|_2^2\rb \E\lb\left\| \bar{X} - \theta\lp P\rp \right\|^2_2 \rb},
\end{align} 
where (a) and (b) follow from the triangular inequality and the Cauchy-Schwartz inequality respectively.
Since $X_i$ and $\theta(P)$ are supported on the unit ball, $\E\lb\left\| \bar{X} - \theta\lp P\rp \right\|^2_2 \rb \asymp 1/n$, so it remains to find a distribution $P^*$ such that 
$$ \min_{\hat{X}} \E\lb \left\| \hat{X} - \theta\lp P^*\rp \right\|_2^2\rb \succeq \frac{d}{n\min\lp \varepsilon^2, \varepsilon,b\rp}.$$
Consider the product Bernoulli model $ Y \sim \prod_{j=1}^d \Ber(\theta_j) $. If we set $\Theta = [1/2-\varepsilon, 1/2+\varepsilon]^d$ for some $\frac{1}{2} > \varepsilon > 0$, then it can be shown that both variance and sub-Gaussian norm of the score function of this model  is $\Theta(1)$ \cite[Corollary~4]{barnes2019lower}. Therefore, applying \cite[Corollary~8]{barnes2019lower} and \cite[Proposition~2, Proposition~4]{barnes2020fisher} yields
$$ \min_{\hat{\theta}} \E\lb \left\| \hat{\theta} - \theta\right\|_2^2\rb \succeq \frac{d^2}{n\min\lp \varepsilon^2, \varepsilon,b\rp}. $$
Finally, if we set $X_i = Y_i/\sqrt{d}$, then each $X_i$ is supported on the unit ball and $\E\lb X_i\rb = \theta/\sqrt{d}$. Therefore
$$ \min_{\hat{X}} \E\lb \left\| \hat{X} - \frac{\theta}{\sqrt{d}} \right\|_2^2\rb \succeq \frac{d}{n\min\lp \varepsilon^2, \varepsilon,b\rp}.$$
Plugging into \eqref{eq:mean_est_lower}, as long as $\min(\varepsilon^2, \varepsilon, k) = o(d)$, the first term dominates and we get the desired lower bound.
\qedwhite

{
\section{Proof of Theorem~\ref{thm:sme}}\label{sec:sme_proof}
The lower bounds follow directly from \cite{bhowmick2018protection} (under $\varepsilon$-LDP constraint) and \cite{an2016distributed} (under $b$-bit communication constraint). For the achievability part, we apply SQKR except that replacing the random sampling step by deterministic grouping.

Let $X_i \diid P$ with $P$ supported on $\mcal{B}(\bm{0},1)$. First, as in the proof of Theorem~\ref{thm:emp_dist_estimation}, by Lemma~\ref{lemma:up_ Kashin} we can write $X_i = \sum_{j=1}^N A_{ij}u_j$ with $N = c_0 d$ and $\lba A_{ij} \rba \leq K/\sqrt{d}, K = \Theta\lp 1\rp$. Since $X_i \diid P$, if we denote $A_i = \lb A_{i1},...,A_{iN} \rb$, then $A_i \diid Q$ for some $Q$ supported on $\lb -\frac{K}{\sqrt{d}}, \frac{K}{\sqrt{d}} \rb^N$.

Now we group $n$ clients into $m \eqDef N/b^*$ groups $\mcal{G}_1,...,\mcal{G}_m$, each with $n b^*/N$ clients, where $b^* \eqDef \min\lp \lceil \varepsilon \log_2 e \rceil, b \rp$. Also, we divide all of $N$ coordinates (of $A_i$) into $m$ groups $\mcal{I}_1,...,\mcal{I}_m$, and each group of clients are responsible for estimating the corresponding group of coordinates of $\theta\lp Q\rp \in \lb -\frac{K}{\sqrt{d}}, \frac{K}{\sqrt{d}}\rb^N$, where $\theta\lp Q\rp = \E_Q[A]$ is the mean of $Q$ and $\theta\lp Q\rp$.

\paragraph{Quantization}
If client $i$ belongs to $\mcal{G}_l$, then it quantizes $A_{ij}$ to $Q_{ij}$ according to 
\begin{equation}
      Q_{ij} \eqDef 
    \begin{cases}
    -\frac{K}{\sqrt{d}}, \text{ with probability } \frac{K/\sqrt{d} - A_{ij}}{2K/\sqrt{d}}, \text{ if } j\in\mcal{I}_l,\\
    \frac{K}{\sqrt{d}}, \text{ with probability } \frac{A_{ij} + K/\sqrt{d}}{2K/\sqrt{d}},  \text{ if } j\in\mcal{I}_l,\\
    0, \text{ else.}
    \end{cases}
\end{equation}
Conditioned on $A_{i}$, $\lbp {Q}_{ij} \mid j \in \mcal{I}_l \rbp$ yields an unbiased estimator of $\lbp A_{ij} \mid j \in \mcal{I}_l \rbp$ and can be described by $\lba \mcal{I}_l \rba = b^*$ bits.

\paragraph{Privatization}
Client $i$ then perturbs the $b^*$-bit message $\lbp {Q}_{ij} \mid j \in \mcal{I}_l \rbp$ into $\lbp \hat{Q}_{ij} \mid j \in \mcal{I}_l \rbp$ via $2^{b^*}$-RR, as described in \eqref{eq:me_privatize}. Similarly, $$\lbp \lp\frac{e^\varepsilon+2^{b^*}-1}{e^\varepsilon-1} \rp\hat{Q}_{ij} \mid j \in \mcal{I}_l \rbp$$ yields an unbiased estimator on $\lbp A_{ij} \mid j \in \mcal{I}_l \rbp$.

\paragraph{Estimation and the $\ell_2$ error}
For all $j \in \mcal{I}_l$, $\hat{A}_{ij} \eqDef \lp\frac{e^\varepsilon+2^{b^*}-1}{e^\varepsilon-1} \rp\hat{Q}_{ij}$ yields an unbiased estimator on $\E_Q\lb A_{ij}\rb$, and note that $\hat{Q}_{ij}\in \lb -\frac{K}{\sqrt{d}}, \frac{K}{\sqrt{d}}\rb$, so the variance of $\hat{A}_{ij}$ is controlled by
$$ \E_Q\lb \lp \hat{A}_{ij} - \theta\lp Q\rp(j) \rp \rb \leq \lp\frac{e^\varepsilon+2^{b^*}-1}{e^\varepsilon-1} \rp^2 \lp\frac{2K}{\sqrt{d}}\rp^2 = O\lp \frac{1}{d\min\lp 1, \varepsilon^2\rp} \rp. $$
Since for each coordinate $j \in \mcal{I}_l$, there are $\lba \mcal{G}_{l}\rba$ clients (samples) that output independent and unbiased estimators $\hat{A}_{ij}$, the estimator
$$ \hat{A}_j \eqDef \frac{1}{\lba \mcal{G}_{l}\rba}\sum_{i\in\mcal{G}_l}\hat{A}_{ij} $$
has variance $$ O\lp \frac{1}{d \lba \mcal{G}_l \rba} \rp = O\lp \frac{1}{n \min\lp b^*, \varepsilon^2\rp} \rp. $$
Therefore, we arrive at
$$ \E\lb \sum_{j=1}^N\lp\hat{A}_j - \E_Q\lb A_j \rb\rp^2 \rb =  O\lp \frac{d}{n \min\lp b^*, \varepsilon^2\rp} \rp. $$
Write $\hat{\theta} = \sum_{j=1}^N \hat{A}_j u_j$ and note that $\theta\lp P \rp = \sum_{j=1}^N \E_Q\lb\hat{A}_j\rb u_j$, so by \eqref{eq:mse_bdd} we conclude that
$$ \E_P\lb \lVert \hat{\theta} - \theta(P) \rVert^2_2 \rb = O\lp \frac{d}{n\min\lp b^*, \varepsilon^2\rp}\rp = O\lp \frac{d}{n\min\lp \varepsilon, \varepsilon^2, b \rp}\rp. $$ \qedwhite
}

%% file: sec_freq_estimation_app.tex
\subsection{Achieving optimal $\ell_1$ and $\ell_2$ error (part~(i) of Theorem~\ref{thm:emp_dist_estimation})} \label{sec:proof_freq_est_achievability}
In this section, we show that Recursive Hadamard Response (RHR) achieves optimal $\ell_1$ and $\ell_2$ estimation error.
\paragraph{Decomposition of Hadamard matrix}
Let us set $B = d/2^{k-1}$. Since $H_d = H_{2^{k-1}} \otimes H_{B}$, for any $j \in [B]$ and $m \in [2^{k-1}]$, if $j' = (m-1)B+j$ (and thus $j \equiv j' \pmod{B} $), we must have
$ \lp H_{d}\rp_{j'} = \lp H_{2^{k-1}} \rp_m \otimes \lp H_b \rp_j$, where $\otimes$ is the Kronecker product.
This allows us to decompose the $j'$-th component of $H_d \cdot X_i$ into
\begin{align}\label{eq:hadmard_decomp}
(H_d)_{j'}\cdot X_i 
     =  \lp \lp H_{2^{k-1}}\rp_m \otimes (H_B)_j\rp\cdot X_i 
    & = \sum_{l = 1}^{2^{k-1}} \lp H_{2^{k-1}}\rp_{m, l}  (H_B)_j\cdot X_i^{(l)},
\end{align}
where $X_i^{l}$ is the $l$-th block of $X_i$, i.e. $X_i^{(l)} \eqDef X_i[(l-1)B+1: l B]$.
Therefore, as long as we know $(H_B)_j\cdot X_i^{(l)}$ for $l = 1,...,2^{k-1}$, we can reconstruct
$(H_d)_{j'}\cdot X_i$, for all $j' \equiv j \pmod{B}$. 

\paragraph{Encoding mechanism}
Let $r_i \sim \text{Uniform}(B)$ be generated from the shared randomness, and consider the following quantizer
$$ Q(X_i, r_i) = \lp (H_B)_{r_i}\cdot X_i^{(l)}\rp_{l=1,...,2^{k-1}} \in \{-1,0,1\}^{2^{k-1}}.$$
Since $X_i$ is one-hot encoded, there is exactly one non-zero $X_i^{(l)}$, so $Q(X_i, r_i)$ can be described by a $k$-bit string (with $k-1$ bits indicating the location of the non-zero entry and $1$ bit indicating  its sign). 

Given $Q(X_i, r_i)$, by \eqref{eq:hadmard_decomp} we can recover $2^{k-1}$ coordinates of $Y_i= H_d\cdot X_i$:
\begin{equation}\label{eq:Y_i}
    Y_i(r') = \lp H_d \rp_{r'}\cdot X_i = \sum_{l = 1}^{2^{k-1}} \lp H_{2^{k-1}}\rp_{m,l} (H_B)_{r_i}\cdot X_i^{(l)} = \lp H_{2^{k-1}} \rp_m\cdot Q(X_i, r_i),
\end{equation} 
for any $r' = (m-1)B+r_i$. Therefore, if we define
\begin{equation}\label{eq:hat_Y_i}
    \hat{Y}_i(Q(X_i, r_i), r_i) \eqDef 
\begin{cases}
\frac{1}{2^{k-1}}Y_i(r'), \text{ if } r' \equiv r_i\\
0, \text{ else,}
\end{cases}
\end{equation} 
then $\E \lb \hat{Y}_i \rb = \frac{1}{d}H_d \cdot X_i$,
where the expectation is taken with respect to $r_i$.

To protect privacy, client $i$ then perturbs $Q(X_i, r_i)$ via $2^k$-RR scheme, since $Q$ takes values on an alphabet of size $2^k$, denoted by $\mathcal{Q}=\{\pm e_1,\dots, \pm e_{2^{k-1}}\}$,
$$ \tilde{Q}_i = 
\begin{cases}
    Q(X_i, r_i), \text{ w.p. } \frac{e^\varepsilon}{e^\varepsilon + 2^k-1}\\
    Q' \in \mathcal{Q} \setminus \lbp Q(X_i, r_i) \rbp, \text{ w.p. } \frac{1}{e^\varepsilon + 2^k-1},
\end{cases}$$
where $\bm{e}_l$ denotes the $l$-th coordinate vector in $\mbb{R}^{2^{k-1}}$.

Client $i$ then sends the $k$-bit report $\tilde{Q}_i$ to the server, and with $\tilde{Q}_i$, the server can compute an estimate of $Q_i$ since
$ \E \lb \tilde{Q}_i \Big| Q(X_i, r_i) \rb = \frac{e^\varepsilon - 1}{e^\varepsilon + 2^k -1} Q(X_i, r_i).$

\paragraph{Constructing estimator for $\hat{D}$}
For a given $\tilde{Q}_i$, we estimate $Y_i$ by $\hat{Y}_i\lp \frac{e^\varepsilon + 2^k -1}{e^\varepsilon -1}\tilde{Q}_i, r_i \rp$,
where $\hat{Y}_i$ is given by \eqref{eq:Y_i} and \eqref{eq:hat_Y_i}, with $Q(X_i, r_i)$ in \eqref{eq:Y_i} replaced by $\tilde{Q}_i$. 
\begin{claim}\label{clm:y_unbiased}
$\hat{Y}_i$ is an unbiased estimator of $Y_i$.
\end{claim}

The final estimator of $D_{X^n} = \frac{1}{n}\sum X_i$ is given by
\begin{equation} \label{eq:hat_D}
 \hat{D}\lp \lp \tilde{Q}_i, r_i \rp_{i=1, ...,n} \rp \eqDef \frac{1}{n}\sum_{i=1}^n H_d\cdot \hat{Y}_i\lp \frac{e^\varepsilon + 2^k -1}{e^\varepsilon -1}\tilde{Q}_i, r_i \rp.  
\end{equation}
Note that by Claim~\ref{clm:y_unbiased}, $\hat{D}$ is an unbiased estimator for $D_{X^n}$. Finally picking $k = \min\lp b,  \lceil \varepsilon \log_2 e \rceil, \lfloor \log d \rfloor \rp$ yields the following bounds.
\begin{claim}\label{clm:bdd_err_emp}
The estimator $\hat{D}$ in \eqref{eq:hat_D} achieves the optimal $\ell_1$ and $\ell_2$ errors:
$$ \E\lb \left\| \hat{D} - D_{X^n} \right\|^2_2 \rb \preceq \frac{d}{n\lp \min{\lbp e^\varepsilon, \lp e^{\varepsilon} -1 \rp^2, 2^b, d\rbp} \rp}\,\,\,\,\, \text{  and} $$
$$ \E \lb \left\| \hat{D} - D_{X^n} \right\|_1 \rb \preceq \frac{d}{\sqrt{n\lp \min{\lbp e^\varepsilon, \lp e^{\varepsilon} -1 \rp^2, 2^b, d\rbp} \rp}}. $$
\end{claim}

This establishes the achievability part of Theorem~\ref{thm:emp_dist_estimation}.\qedwhite

\subsection{Algorithms}\label{subsec:freq_est_alg}
We summarize our proposed scheme  RHR scheme below:

\begin{algorithm}[H]\label{alg:emp_encode}
\SetAlgoLined
\KwIn{client index $i$, observation $X_i$, privacy level $\varepsilon$, alphabet size $d$}
\KwResult{ Encoded message $\lp \tilde{\texttt{sign}}, \tilde{\texttt{loc}} \rp$ }
 Set $D = 2^{\lceil\log d\rceil}$, 
 $k = \min \lp b, \lceil \varepsilon \log_2 e \rceil \rp$, $B = D/2^{k-1}$\;
 Draw $r_i$ from $\text{uniform}(B)$ using public-coin \;
 \Begin{
 $\texttt{loc} \la \lceil \frac{X_i}{B} \rceil$\; 
 $\texttt{sign} \la \lp H_d\rp _{r_i,X_i}$\; 
 $ \lp \tilde{\texttt{sign}}, \tilde{\texttt{loc}} \rp\la 2^{k}-\text{RR}_{\varepsilon} \lp \lp \texttt{sign}, \texttt{loc} \rp \rp$
 \tcc*[f]{ $\lp \text{sign}, \text{loc} \rp$ as a $k$-bit string}\;
 }
 \caption{Encoding mechanism $\tilde{Q}_i$ (at each client)}
\end{algorithm}
Notice that computing any entry of $H_d$ takes $O\lp \log d\rp$ Boolean operations, and uniformly sampling a $k$-bit string takes $O(k)$ time. Therefore the computation cost at each client is $O\lp \log d \rp$ time. Also note that the encoded message is a $k$-bit binary string, and therefore the communication cost at each client is $k = \min\lp b, \lceil \varepsilon \log_2(e) \rceil \rp \leq b$.

Once receiving the $k$-bit messages from all clients, the server does the following operation:

\begin{algorithm}[H]
\SetAlgoLined
\KwIn{ $( \tilde{\texttt{sign}}[1:n], \tilde{\texttt{loc}}[1:n] )$, privacy level $\varepsilon$, alphabet size $d$}
\KwResult{ $\hat{D}$ }
 Set $D = 2^{\lceil\log d\rceil}$, $k = \min \lp b, \lceil \varepsilon \log_2 e \rceil \rp$, $B = D/2^{k-1}$\;
 Partition messages into groups $\mcal{G}_1,...,\mcal{G}_B$, with message $i$ in $\mcal{G}_{r_i}$\;
 \ForAll{$j = 1,...,B$}{
 $\mcal{G}_j^+ \la \lbp \tilde{\texttt{loc}}(i) \,|\, i \in \mcal{G}_j, \tilde{\texttt{sign}}(i) = +1 \rbp$\;
 $\mcal{G}_j^- \la \lbp \tilde{\texttt{loc}}(i) \,|\, i \in \mcal{G}_j, \tilde{\texttt{sign}}(i) = -1 \rbp$\;
 $\msf{Emp}_j \la \lp\text{empirical distribution}(\mcal{G}_j^+) - \text{empirical distribution}(\mcal{G}_j^-)\rp \cdot \frac{e^\varepsilon + 2^{k}-1}{e^\varepsilon -1}$\;
 \ForAll{$l = 0,...,2^{k-1}-1$}{
 $\hat{E}[l\cdot B+j]\la \msf{FWHT}(\msf{Emp}_j)[l]$  \tcc*[f]{fast Walsh-Hadamard transform}
 }
 }
$\hat{D} \la \frac{1}{d}\cdot\msf{FWHT}\lp \hat{E}\rp$\;
\caption{Estimator of $D_{X^n}$ (at the server)}
\end{algorithm}

Partitioning $n$ samples into $B$ groups and computing the empirical distribution of each group takes $O(n)$ time, and the fast Walsh-Hadamard transform can be implemented in $O\lp d \log d \rp$ time. Hence the decoding complexity is $O\lp n + d\log d \rp$.

\subsection{Lower Bound on $\ell_1$ and $\ell_2$ errors in Theorem~\ref{thm:emp_dist_estimation}}
We can bound the error by considering the worst case Bayesian setting, i.e. by imposing a prior distribution $\bm{p}$ on $X_1,..., X_n$ and applying the converse part of Theorem~\ref{thm:dist_estimation} in Section~\ref{sec:dist_est}. 

Let $X_1,...,X_n \diid \bm{p}$. Then for any $\hat{D}(X^n)$, we must have
\begin{align}\label{eq:emp_lower_bdd}
    \max_{X^n \sim \bm{p}}\E\lb \left\| \hat{D} - D_{X^n} \right\|_2^2 \rb 
    & \overset{\text{(a)}}{\geq} \max_{\bm{p}}\E\lb \lp \left\| \hat{D} - \bm{p} \right\|_2 -  \left\| D_{X^n} - \bm{p} \right\|_2\rp^2\rb \nonumber\\
    & \geq \max_{\bm{p}}\lp \E\lb\left\| \hat{D} - \bm{p} \right\|_2^2\rb - 2\E\lb\left\|\hat{D} - \bm{p}\right\|_2 \left\| D_{X^n} - \bm{p}\right\|_2\rb\rp \nonumber \\
    & \overset{\text{(b)}}{\geq} \max_{\bm{p}}\lp \E\lb\left\| \hat{D} - \bm{p} \right\|_2^2\rb - 2\sqrt{\E\lb\left\|\hat{D} - \bm{p}\right\|_2^2 \rb \E\lb \left\| D_{X^n} - \bm{p}\right\|_2^2\rb}\rp
\end{align}
where (a) and (b) follow from the triangular inequality and the Cauchy-Schwarz inequality respectively. By Theorem~\ref{thm:dist_estimation}, there exists a worst case $\bm{p}^*$ such that
\begin{equation}\label{eq:D_hat_p_bdd}
    c\frac{d}{n} \lp \frac{1}{\min{\lbp e^\varepsilon, \lp e^{\varepsilon} -1 \rp^2, 2^b\rbp}} \rp \leq \E\lb  \left\| \hat{D} - \bm{p}^* \right\|_2^2\rb \leq C\frac{d}{n} \lp \frac{1}{\min{\lbp e^\varepsilon, \lp e^{\varepsilon} -1 \rp^2, 2^b\rbp}} \rp,
\end{equation}
 
for some constants $c$ and $C$.
On the other hand, the $\ell_2$ convergence of $D(X^n)$ to $\bm{p}$ is $O\lp1/n\rp$ for any $\bm{p}$, which gives us
\begin{equation}\label{eq:D_p_bdd}
     \E\lb \left\| D_{X^n} - \bm{p}^*\right\|_2^2\rb \leq c'\frac{1}{n}.
\end{equation}

Plugging \eqref{eq:D_hat_p_bdd} and \eqref{eq:D_p_bdd} back into \eqref{eq:emp_lower_bdd} yields
\begin{align*}
    &\max_{X^n \sim \bm{p}}\E\lb \left\| \hat{D} - D_{X^n} \right\|_2^2 \rb \\
    &\geq C_1\frac{d}{n} \lp \frac{1}{\min{\lbp e^\varepsilon, \lp e^{\varepsilon} -1 \rp^2, 2^b\rbp}} \rp - C_2
    \frac{1}{n} \sqrt{\frac{d}{\min{\lbp e^\varepsilon, \lp e^{\varepsilon} -1 \rp^2, 2^b\rbp}}}.
\end{align*}
Thus as long as $\min\lp e^\varepsilon, \lp e^\varepsilon -1 \rp^2, 2^b  \rp = o(d)$, the first term dominates and the desired $\ell_2$ lower bound follows.

For the case of $\ell_1$, we similarly have
\begin{align}\label{eq:emp_lower_bdd_l1}
    \max_{X^n \sim \bm{p}}\E\lb \left\| \hat{D} - D_{X^n} \right\|_1 \rb 
    & \geq \max_{\bm{p}}\lp \E\lb \left\| \hat{D} - \bm{p} \right\|_1 \rb  - \E\lb \left\| D_{X^n} - \bm{p} \right\|_1\rb\rp
\end{align}
It is well-known that $\E\lb \left\| D(X^n) - \bm{p} \right\|_1\rb \leq \sqrt{d/n}$ (for instance, see \cite{han2015minimax}), and by the converse part of Theorem~\ref{thm:dist_estimation}
$$ \max_{\bm{p}}\E\lb \left\| \hat{D} - \bm{p} \right\|_1 \rb \geq \sqrt{\frac{d^2}{n \min{\lbp e^\varepsilon, \lp e^{\varepsilon} -1 \rp^2, 2^b\rbp}}}.$$
Plugging this into \eqref{eq:emp_lower_bdd_l1} yields the $\ell_1$ lower bound.
\qedwhite

\subsection{Achieving optimal $\ell_\infty$ error (part~(ii) of Theorem~\ref{thm:emp_dist_estimation} )}\label{proof:freq_l_infty}

To obtain an upper bound on $\ell_\infty$ error, we extend the \texttt{TreeHist} protocol in \cite{Bassily2017}, a $1$-bit LDP heavy hitter estimation mechanism, to communicate $b$ bits  and satisfy a desired privacy level $\varepsilon$. A simpler version of \texttt{TreeHist} protocol, which is not optimized for computational complexity, is as follows: we first perform Hadamard transform on $X_i$, and sample one random coordinate with public randomness $r_i$. The $1$-bit message is then passed through a binary $\varepsilon$-LDP mechanism. We can show that from the perturbed outcomes, the server can construct an unbiased estimator of $X_i$ with bounded sub-Gaussian norm, and the $\ell_\infty$ error will be $O(\sqrt{\log d/ n \varepsilon^2})$.

To extend this scheme to an arbitrary privacy regime and an arbitrary communication budget of $b$ bits, we independently and uniformly sample the Hadamard transform of $X_i$ for $k = \min\lp b, \lceil \varepsilon \rceil \rp$ times. Each $1$-bit sample is then perturbed via a $\varepsilon'$-LDP mechanism with $\varepsilon' \eqDef \varepsilon / k$. 

Note that under the distribution-free setting, the randomness comes only from the sampling  and the privatization steps, so we could view each re-sampled and perturbed message as generated from a fresh new copy of $X_i$ since $X_i$ is not random. Equivalently, this boils down to a frequency estimation problem with $n' = nk$ clients and under $\varepsilon' = \varepsilon /k$ and gives us the $\ell_\infty$ error
$$O\lp \sqrt{\frac{\log d}{n'\lp \varepsilon'\rp^2}} \rp= O\lp \sqrt{\frac{\log d}{n\min\lp \varepsilon^2, \varepsilon, b\rp}} \rp.$$
Below we describe the details.

\paragraph{Encoding mechanism} Set $k = \min\lp b, \lceil \varepsilon \rceil \rp$. For each $X_i$, we randomly sample $\lp H_d\rp_{X_i}$ (i.e. the $X_i$-th column of $H_d$) $k$ times, identically and independently by using the shared randomness. Let $r^{(1)}_i,...,r^{(k)}_i$ be the sampled coordinates, which are known to both the server and node $i$, and $\lp H_d\rp_{X_i, r^{(\ell)}_i}$ be the sampling outcomes.
Then due to the orthogonality of $H_d$, for all $j\in[d], \ell\in[k]$, 
\begin{equation}\label{eq:H_unbias}
    \E\lb (H_d)_{j, r^{(\ell)}_i}\cdot (H_d)_{X_i, r^{(\ell)}_i}\rb = 
\begin{cases}
1, \text{ if } j = X_i\\
0, \text{ if } j \neq X_i,
\end{cases}
\end{equation}
where the expectation is taken over $r_i^{(\ell)}$.

We then pass $\lbp(H_d)_{X_i, r^{(\ell)}_i} \Big| \ell = 1,...,k \rbp$ through $k$ binary $\varepsilon'$-LDP channels sequentially, with $\varepsilon' \eqDef \varepsilon/k$. By the composition theorem of differential privacy, the privatized outcomes, denoted as $\lbp \tilde{(H_d)}_{X_i, r_i^{(\ell)}} \rbp$, satisfy $\varepsilon$-LDP.

\paragraph{Estimation of $D_{X^n}$} Observe that
$$ \E \lb \lp \frac{e^{\varepsilon'} + 1}{e^{\varepsilon'} - 1}\rp\tilde{(H_d)}_{X_i, r^{(\ell)}_i} \mv (H_d)_{X_i, r^{(\ell)}_i} \rb = (H_d)_{X_i, r^{(\ell)}_i}, $$
where the expectation is with respect to the privatization.
Therefore 
$$ \hat{X}^{(\ell)}_i(j) \eqDef \lp \frac{e^{\varepsilon'} + 1}{e^{\varepsilon'} - 1}\rp(H_d)_{j, X_i}\tilde{(H_d)}_{X_i, r^{(\ell)}_i}$$
defines an unbiased estimator of $X_i(j)$. Moreover, 
$$ \lba \hat{X}^{(\ell)}_i(j) - X_i(j) \rba \leq \lp \frac{e^{\varepsilon'} + 1}{e^{\varepsilon'} - 1} +1\rp \text{ a.s.}, $$
so $\hat{X}^{(\ell)}_i(j)$ has sub-Gaussian norm bounded by 
\begin{equation}\label{eq:sigma}
\sigma \leq 2\frac{e^{\varepsilon'} + 1}{e^{\varepsilon'} - 1}.
\end{equation}
Finally, we estimate $D_{X^n}(j)$ by
$$ \hat{D}(j) = \frac{1}{n k}\sum_{i = 1}^n \sum_{\ell = 1}^{k} \hat{X}^{(\ell)}_i(j).  $$
Observe that
\begin{align}
    \hat{D}(j) - D_{X^n}(j) = \frac{1}{n k}\sum_{i = 1}^n \sum_{\ell = 1}^{k} \lp \hat{X}^{(\ell)}_i(j) - X_i(j)\rp
\end{align}
has sub-Gaussian norm bounded by $ \sigma/\sqrt{nk}$, where $\sigma$ is given by \eqref{eq:sigma}.

To bound the $\ell_\infty$ norm, we apply the maximum bound (see, for instance, \cite[Chapter~2]{wainwright2019high}) for sub-Gaussian random variables (note that for $j, j'$, $\hat{D}(j)$ and $\hat{D}(j')$ are not independent):
\begin{align}\label{eq:bdd_ell_infty}
    \E \lb \max_{j \in [d]} \lba \hat{D}(j) - D_{X^n}(j)\rba \rb 
     \leq 2\sqrt{\sigma^2 \log d} 
     = 4\sqrt{ \lp \frac{e^{\varepsilon'} + 1}{e^{\varepsilon'} - 1} \rp^2 \frac{\log d}{n k}}
    & \overset{\text{(a)}}{\asymp} \sqrt{\frac{\log d}{n \min\lp\varepsilon, \varepsilon^2, k\rp}},
\end{align}
where (a) holds since if $\varepsilon = o(1)$, then $k = 1$ and hence
    $$ \lp \frac{e^{\varepsilon'} + 1}{e^{\varepsilon'} - 1} \rp^2 \asymp \frac{1}{\varepsilon^2}; $$
otherwise $\varepsilon = \Omega(1)$ and $\varepsilon' = \Omega(1)$, so 
    $$ \lp \frac{e^{\varepsilon'} + 1}{e^{\varepsilon'} - 1} \rp^2 \asymp 1. $$
Both cases are upper bounded by \eqref{eq:bdd_ell_infty}, so the result follows. \qedwhite

\begin{remark}
Notice that in the high privacy regime $\varepsilon = o(1)$, the upper bound matches the lower bound in \cite{Bassily2015}. For general privacy regimes with limited communication, however,  we do not know whether the upper bound is tight or not. This remains as an open question.
\end{remark}

\section{Proof of Theorem~\ref{thm:dist_estimation}}\label{sec:dist_est_app}
The construction of the distribution estimation scheme mainly follows Section~\ref{sec:proof_freq_est_achievability}, except we replace the random sampling step by a deterministic grouping idea. We will use the same notation as in Section~\ref{sec:proof_freq_est_achievability}.

\paragraph{Encoding mechanism}
We group $n$ samples into $B$ equal-sized groups, each with $n' = n/B$ samples. For sample $X_i \in \mcal{G}_j$, we quantize it to a $2^{k-1}$-dimensional $\{1, 0, -1\}$ vector:
$$ Q_j(X_i) =
\begin{bmatrix}
 (H_B)_j\cdot X_i^{(1)}  \\
(H_B)_j\cdot X_i^{(2)}\\
\vdots\\
(H_B)_j\cdot X_i^{(2^{k-1})}
\end{bmatrix}
\in \{-1,0,1\}^{2^{k-1}}. $$
Since $X_i$ is one-hot encoded, there is only one $l \in \{1, ..., 2^{k-1}\}$ such that $ (H_B)_j\cdot X_i^{(l)}\neq 0$, so $Q_j(X_i)$ can be described by $k$ bits ($1$ bit for the sign and $(k-1)$ bits for the location of the non-zero element). Also notice that 
$$ \E \lb Q_j(X_i) \rb = 
\begin{bmatrix} 
 (H_B)_j\cdot \bm{p}^{(1)} \\ 
(H_B)_j\cdot \bm{p}^{(2)} \\
\vdots \\
(H_B)_j \cdot \bm{p}^{(2^{k-1})} 
\end{bmatrix}, $$
where $\bm{p}^{(l)} \eqDef \bm{p}[(l-1)B+1:l B]$. By \eqref{eq:hadmard_decomp}, the estimator $\hat{q}_{j'} = \lan \lp H_{2^{k-1}} \rp_m, Q_j(X_i) \ran$ is unbiased for $q_{j'}$ (where $j' = (m-1)B+j$).

We further perturb $Q_j$ via $2^k$-RR scheme, since $Q$ takes values on an alphabet of size $2^k$, denoted by $\mathcal{Q}=\{\pm e_1,\dots, \pm e_{2^{k-1}}\}$,
$$ \tilde{Q}_j = 
\begin{cases}
    Q_j, \text{ w.p. } \frac{e^\varepsilon}{e^\varepsilon + 2^k-1}\\
    Q' \in \mathcal{Q} \setminus \lbp Q_j \rbp, \text{ w.p. } \frac{1}{e^\varepsilon + 2^k-1},
\end{cases}$$
where $\bm{e}_l$ denotes the $l$-th coordinate vector in $\mbb{R}^{2^{k-1}}$.
This gives us 
$$ \E \lb \tilde{Q}_j \rb = \frac{e^\varepsilon - 1}{e^\varepsilon + 2^k -1} \E \lb Q_j\rb. $$ 
Therefore $\frac{e^\varepsilon+2^k-1}{e^\varepsilon -1} \tilde{Q}_j$ yields an unbiased estimator of 
$$
\begin{bmatrix} 
(H_B)_j\cdot \bm{p}^{(1)} \\ 
(H_B)_j\cdot \bm{p}^{(2)}  \\
\vdots \\
(H_B)_j\cdot \bm{p}^{(2^{k-1})}
\end{bmatrix}. $$

\paragraph{Constructing the estimator for $\bm{p}$}
For each $j' \equiv j \pmod B$, we estimate $\lp H_{2^{k-1}} \rp_m\cdot Q_j(X_i), i \in \mcal{G}_j$ (recall that $j' = j+(m-1)B$). 
Define the estimator
\begin{align*}
    \hat{q}_{j'}\lp \lbp X_i, i \in \mcal{G}_j \rbp \rp
    & = \frac{1}{\lba \mcal{G}_j \rba}\sum_{i \in \mcal{G}_j}\lp H_{2^{k-1}} \rp_m\cdot \lp \frac{e^\varepsilon+2^k-1}{e^\varepsilon -1}\rp \tilde{Q}_j(X_i)\\
    & = \frac{B}{n}\lp \frac{e^\varepsilon+2^k-1}{e^\varepsilon -1}\rp\sum_{i \in \mcal{G}_j}\lp H_{2^{k-1}} \rp_m \tilde{Q}_j(X_i).
\end{align*}

The MSE of $\hat{q}_{i'}$ can be obtained by
\begin{align}\label{eq:bdd_q_var_2}
    \E \lb \lp \hat{q}_{j'} - q_{j'} \rp^2\rb 
    &\overset{\text{(a)}}{=} \Var\lp \hat{q}_{i'} \rp \nonumber\\
    &\overset{\text{(b)}}{=} \frac{d}{n 2^{k-1}}\lp \frac{e^\varepsilon+2^k-1}{e^\varepsilon -1}\rp^2 \Var\lp  \lp H_{2^{k-1}} \rp_m\cdot \tilde{Q}_j(X_i) \rp \nonumber\\
    &\overset{\text{(c)}}{\leq} \frac{d}{n 2^{k-1}}\lp \frac{e^\varepsilon+2^k-1}{e^\varepsilon -1}\rp^2,
\end{align}
where (a) is due to the unbiasedness of $\hat{q}_{j'}$, (b) is due to the independence across $X_i$, and (c) is because $\lan \lp H_{2^{k-1}} \rp_m, \tilde{Q}_j\ran$ only takes value in $\{-1, 1\}$.

Finally, let $\hat{\bm{p}}$ be the inverse Hadamard transform of $\hat{\bm{q}}$, the MSE is
\begin{align*}
    \E \left\| \hat{\bm{p}} - \bm{p} \right\|^2_2
    & = \E \lb \lan \hat{\bm{p}} - \bm{p}, \hat{\bm{p}} - \bm{p} \ran \rb \\
    & = \E  \lb \lp \hat{\bm{q}} - \bm{q} \rp^\intercal \lp H_d^{-1}\rp ^\intercal H_d^{-1} \lp \hat{\bm{q}} - \bm{q} \rp \rb \\
    & = \frac{1}{d}\E \left\| \hat{\bm{q}} - \bm{q} \right\|^2_2 \\
    & \leq \frac{d}{n 2^k}\lp \frac{e^\varepsilon+2^k-1}{e^\varepsilon -1}\rp^2 \\
    & = O\lp \frac{d}{n 2^k} \lp \frac{e^\varepsilon +2^k}{e^\varepsilon - 1} \rp^2 \rp,
\end{align*}
where the last inequality holds due to \eqref{eq:bdd_q_var_2}. 

Picking $k = \min\lp b,  \lceil \varepsilon \log_2 e \rceil, \lfloor \log d \rfloor \rp$ yields 
\begin{align*}
    \E \left\| \hat{\bm{p}} - \bm{p} \right\|^2_2 
    & = O\lp \frac{d}{n \min\lp 2^b, e^\varepsilon, d\rp} \lp \frac{e^\varepsilon}{e^\varepsilon - 1} \rp^2 \rp.
\end{align*}
Observe that if $e^\varepsilon = O(2^b)$, then $e^\varepsilon \preceq 2^b$, so 
    $\E \left\| \hat{\bm{p}} - \bm{p} \right\|^2_2 = O\lp \frac{d e^\varepsilon}{n\lp e^\varepsilon - 1\rp^2} \rp.$
On the other hand, if $e^\varepsilon = \Omega(2^b)$, then $\frac{e^\varepsilon}{e^\varepsilon -1} = \theta(1)$, and
    $\E \left\| \hat{\bm{p}} - \bm{p} \right\|^2_2 = O\lp \frac{d}{n \min \lp 2^b, d \rp} \rp.$

Therefore we conclude that 
$$\E \left\| \hat{\bm{p}} - \bm{p} \right\|^2_2 \preceq \max \lp \frac{d}{n \min\lp 2^b, d\rp}, \frac{d e^\varepsilon}{n\lp e^\varepsilon -1\rp^2} \rp \asymp \frac{d}{n} \lp \frac{1}{\min{\lbp e^\varepsilon, \lp e^{\varepsilon} -1 \rp^2, 2^b, d\rbp}} \rp.$$

Finally, by Jensen's inequality and Cauchy-Schwarz inequality, we also have
\begin{align*}
    \E \lb \left\| \hat{\bm{p}} - \bm{p} \right\|_1 \rb
    \leq \lp \E\lb\left\| \hat{\bm{p}} - \bm{p} \right\|_1^2\rb\rp^{\frac{1}{2}}
    \leq \lp d\cdot\E\left\| \hat{\bm{p}} - \bm{p} \right\|_2^2\rp^{\frac{1}{2}}
    \preceq \frac{d}{\sqrt{n\lp \min{\lbp e^\varepsilon, \lp e^{\varepsilon} -1 \rp^2, 2^b, d\rbp} \rp}},
\end{align*}
establishing the achievability part of Theorem~\ref{thm:dist_estimation}.\qedwhite

\subsection{Algorithms and analysis}
Each client runs the following algorithm:

\begin{algorithm}[H]
\SetAlgoLined
\KwIn{client index $i$, observation $X_i$, privacy level $\varepsilon$, alphabet size $d$}
\KwResult{ Encoded message $\lp \tilde{\texttt{sign}}, \tilde{\texttt{loc}} \rp$ }
 Set $D = 2^{\lceil\log d\rceil}$. 
 Set $k = \min \lp b, \lceil \varepsilon \log_2 e \rceil \rp$, $B = D/2^{k-1}$\;
 \Begin{
 $j \la i\mod B$ \tcc*[f]{assign user $i$ to group $j$}\;
 $\texttt{loc} \la \lceil \frac{X_i}{B} \rceil$\;
 $\texttt{sign} \la \lp H_d\rp _{j,X_i}$ \;
 $ \lp \tilde{\texttt{sign}}, \tilde{\texttt{loc}} \rp\la k\text{RR}_{\varepsilon} \lp \lp \texttt{sign}, \texttt{loc} \rp \rp$
 \;
 }
 \caption{Encoding mechanism (at each client)}\label{alg:freq_est_encoding}
\end{algorithm}

As in Algorithm~\ref{alg:emp_encode}, the computation cost at each client is $O\lp \log d \rp$. Also note that the encoded message is a $k$-bit binary string, and therefore the communication cost at each client is $k = \min\lp b, \varepsilon \log_2(e) \rp \leq b$.

Upon receiving the privatized $k$-bit  messages from the clients, the server runs the following algorithm:

\begin{algorithm}[H]
\SetAlgoLined

\KwIn{ $( \tilde{\texttt{sign}}[1:n], \tilde{\texttt{loc}}[1:n] )$, privacy level $\varepsilon$, alphabet size $d$}
\KwResult{ $\bm{\hat{p}}$ }
 Set $D = 2^{\lceil\log d\rceil}$, $k = \min \lp b, \lceil \varepsilon \log_2 e \rceil \rp$, $B = D/2^{k-1}$\;
 Partition messages into groups $\mcal{G}_1,...,\mcal{G}_B$, with message $i$ in $\mcal{G}_j$ if $i\equiv j \pmod{B}$\;

 \ForAll{$j = 1,...,B$}{
 $\mcal{G}_j^+ \la \lbp \tilde{\texttt{loc}}(i) \,|\, i \in \mcal{G}_j, \tilde{\texttt{sign}}(i) = +1 \rbp$\;
 $\mcal{G}_j^- \la \lbp \tilde{\texttt{loc}}(i) \,|\, i \in \mcal{G}_j, \tilde{\texttt{sign}}(i) = -1 \rbp$\;
 $D_j \la \lp\text{empirical distribution}(\mcal{G}_j^+) - \text{empirical distribution}(\mcal{G}_j^-)\rp \cdot \frac{e^\varepsilon + 2^{k}-1}{e^\varepsilon -1}$\;
 \ForAll{$l = 0,...,2^{k-1}-1$}{
 $\hat{\bm{q}}[l\cdot B+j]\la \msf{FWHT}(D_j)[l]$  \;
 }
 }

$\bm{\hat{p}} \la \frac{1}{d}\cdot\msf{FWHT}\lp \hat{\bm{q}}\rp$\;
\caption{Estimation of $\bm{p}$ (at the server)}\label{alg:freq_est_decoding}
\end{algorithm}

Partitioning $n$ samples into $B$ groups and computing the empirical distribution of each group takes $O(n)$ time, and the fast Walsh-Hadamard transform can be performed in $O\lp d \log d \rp$ time. Hence the decoding complexity is $O\lp n + d\log d \rp$.